# Certainty Closure: Reliable Constraint Reasoning with Incomplete or Erroneous Data

NEIL YORKE-SMITH and CARMEN GERVET

IC–Parc, Imperial College London

Constraint Programming (CP) has proved an effective paradigm to model and solve difficult combinatorial satisfaction and optimisation problems from disparate domains. Many such problems arising from the commercial world are permeated by data uncertainty. Existing CP approaches that accommodate uncertainty are less suited to uncertainty arising due to incomplete and erroneous data, because they do not build reliable models and solutions guaranteed to address the user's genuine problem as she perceives it. Other fields such as reliable computation offer combinations of models and associated methods to handle these types of uncertain data, but lack an expressive framework characterising the resolution methodology independently of the model.

We present a unifying framework that extends the CP formalism in both model and solutions, to tackle ill-defined combinatorial problems with incomplete or erroneous data. The *certainty closure framework* brings together modelling and solving methodologies from different fields into the CP paradigm to provide reliable and efficient approches for uncertain constraint problems. We demonstrate the applicability of the framework on a case study in network diagnosis. We define resolution forms that give generic templates, and their associated operational semantics, to derive practical solution methods for reliable solutions.

Categories and Subject Descriptors: I.2.3 [**ARTIFICIAL INTELLIGENCE**]: Deduction and Theorem Proving—*Uncertainty, logic and constraint programming*

## 1. INTRODUCTION

Data uncertainties are inherent in the real world. They permeate many commercial planning and resource management problems that can be cast as constraint satisfaction and optimisation problems. Forms of data uncertainty are for instance:

(1) future events and changes (scheduling [Davenport and Beck 2000], restaurant management [Vidotto et al. 2007])

(2) stochastic demand, cancellations (hotel reservation [Benoist et al. 2001])

(3) stochastic requests, anticipated changes (oil platform supply [Fowler and Brown 2000], vehicle routing [Bent and Hentenryck 2004])

(4) inadequate demand profiles due to market privatisation (energy portfolio management [Gervet et al. 1999])

(5) partial information, measurement errors (network optimisation [Medina et al. 2002], scene recognition [Dovier et al. 2005])

The data uncertainty can be due to (1) the dynamic and unpredictable nature of the commercial world (first three examples), but also due to (2) the information available to those modelling the problem (last two examples). In this article we are concerned with the latter form of uncertainty, which can arise when the data is not fully known, and also when it is erroneous. Justification of probabilistic modelling in such problems is often lacking because data trends are obsolete, inexistent, or inappropriate, or because scenario-based reasoning is not the desired methodology.

Our motivation to consider the second form of uncertainty above originated in a Network Traffic Analysis Problem (NTAP). Given a known, fixed network with incomplete and possibly erroneous traffic volume measurements at routers, the objective is to determine the minimum and maximum traffic flow values between any two end-point routers. This information, often stored in traffic flow matrices, is valuable to optimise network usage and routing. However, due to the overwhelming amount of information, in practice the data is partial; and due to practical measurement difficulties (e.g. unrecorded packet loss), the data acquired in the problem is frequently erroneous. The end user's problem must be satisfiable, because the network exists and is executing. For the NTAP, the existing approximation and data correction methods were effective to derive solutions but could not guarantee that the user's problem was being addressed. We sought to diagnose whether this was significant in practice.

To achieve this aim, we considered that the user's perception of the constraint problem is her *genuine* problem. Our aim was to provide a *reliable and tractable* model that guarantees to contain the genuine problem, and then to derive reliable solutions from this model.

Constraint Programming (CP) has proved an effective paradigm to model and solve difficult combinatorial search and optimisation problems from disparate application domains [Wallace 1996]. A combinatorial problem is modelled as a *constraint satisfaction problem* (CSP). The modelling and solving of the problem are separated, allowing both expressive, generic modelling and powerful, specialised solving techniques. Based in the CP paradigm, we defined a new concept of model for the NTAP, and developed an efficient solving method to derive the desired solution. We could then determine whether the existing approaches to the problem were reliable, i.e. whether the solutions obtained are potential solutions to the genuine problem. This experimental work is presented in depth in this paper.

The nature of the solution sought by the user to an ill-defined problem depends on several factors: the nature of the data uncertainty, the modelling approach taken, and the user's requirements for the outcome. In the NTAP, for example, the solution sought, bounds on the end-to-end flows, is the projection of the complete solution set onto each flow.

Research to accommodate this type of uncertainty and solution in CP is still sparse; as in the NTAP, the main existing approaches rely on approximation or error correction. Although stochastic approaches can be expressive, they exhibit fun-damental difficulties because there is no principled basis in such problems to attach probabilities to erroneous data, and if we could it would raise the computational complexity of the solving. By contrast, commonly used in robust control, *convex modelling* is a technique from reliable computation whereby uncertain quantities are tractably enclosed with what is known about them. For example, a measured value is enclosed by an interval. Convex modelling relates to some CP frameworks, notably numerical CSPs but also continuous quantified CSPs. While these frameworks could partially be applied for ill-defined problems, the development of this idea has been limited, especially for discrete data, and the complexity of solving the general quantified CSP problem is an obstacle.

Thus, on the one hand we have existing CP extensions that are inadequate to solve our ill-defined problems (either due to the type of uncertainty in the problems, the nature of the solution sought, or the level of generality of the frameworks); yet the general CP paradigm is suitable to describe uncertainty semantics independently of the underlying solving techniques. On the other hand, fields such as operational research (OR) and reliable computation offer combinations of tractable models and associated methods to handle incomplete and erroneous data uncertainty (e.g. interval structures for real-valued data; other convex forms), but do not have a framework characterising the solving methodology independently of the model and techniques. There is a gap between suitable models and methods and a generic, practical CP framework.

Our objective is to fill this gap, by defining a unifying framework that defines the properties of an extension to the CP formalism for ill-defined problems with incomplete and erroneous data, as well as resolution forms that give generic templates and associated operational semantics to derive efficient solution methods. This framework shows how existing models and methods from different fields can be brought together as a pragmatic means towards reliable reasoning in the presence of incomplete and erroneous data.

This article presents the *certainty closure framework* by proposing a generic CSP model and practical, constraint-based resolution forms to reason about such ill-defined problems effectively. We give instances of the resolution forms that constitute a first step towards an automatic tool for reliable constraint reasoning.

In this article we concentrate mainly on describing resolution forms and algorithms for the complete solution set for ill-defined satisfaction problems. Instances of the resolution forms for other types of solution, and extensions of our model for optimisation problems, can be found in the PhD dissertation in [Yorke-Smith 2004], upon which this work is mainly based.

*Organisation.* In Section 2 we survey related work in CP and other fields. In Section 3 we give the essentials of our formal framework in terms of the *uncertain CSP* and a *closure* as its solution. The application of the framework to the Network Traffic Analysis Problem introduced above forms the extended case study of Section 4. In Section 5 we study instances of the uncertain CSP model and their properties, and set forth efficient generic means to solve the models. We conclude and outline future work in Section 6.

## 2.  RELATED WORK

Ill-defined combinatorial problems can not be formulated using probabilistic models such as stochastic CSP [Tarim et al. 2006] when the necessary data trends are either obsolete (e.g. market privatisation), inexistent, or inadequate (e.g. erroneous measurements). Problems with incomplete or erroneous data often fall into this category (e.g. [Gervet et al. 1999; Jaulin 2006]).

The simplest approach often used in CP to model such ill-defined problems is deterministic. A single value is chosen for each uncertain data item (each *parameter*). This selection might be done by using a data error correction model. Error correction to a best deterministic approximation, while widely used as a pragmatic approach, does not aim to ensure reliable solutions, in the sense that it aims primarily to build a satisfiable model rather than to guarantee that the genuine problem is being addressed. Nonetheless, such a model is often perceived as close enough to the "true" model. One of our objectives in this paper is to derive a CP framework and model to diagnose the reliability of such approaches.

Alternative CP approaches of relevance include *mixed CSP*, *numerical CSP*, and *quantified CSP*. Recall that a CSP is a tuple $\langle \mathcal{V}, \mathcal{D}, \mathcal{C} \rangle$, where $\mathcal{V}$ is a finite set of variables, $\mathcal{D}$ is the set of corresponding variable domains (classically finite), and $\mathcal{C}$ is a finite set of constraints. A solution is a complete consistent value assignment.

Although not specified, the focus of the *mixed CSP* framework [Fargier et al. 1994] leans towards uncertainty due to incomplete knowledge or unobserved future. The constraint model of a mixed CSP is restricted to the discrete case. The outcome sought (when no further knowledge will be obtained about the parameter values) is a *robust solution*: one solution that satisfies all constraints under as many *realisations* of the data (possible worlds) as simultaneously possible. The mixed CSP framework introduces the idea of controllability into CP, by distinguishing decision and non-decision variables. The first type of variables are *controllable* by the agent: their values may be chosen. The second type, known in operational research as *parameters*, are *uncontrollable*: their values are assigned by 'Nature', i.e. extrogeneous factors. The model adds expressiveness to a classical CSP, but is generally not suitable to our ill-defined problems because it does not consider also continuous data, and does not consider different types of solutions beyond a robust solution. In the case of erroneous data, one is certainly not looking for a solution that satisfies as many erroneous models as possible.

*Numerical* or *interval CSPs* (NCSPs) extend the classical CSP with continuous variables. In principle they can be used to approximate and reason with continuous (but less readily discrete) uncertain data represented by intervals. We find the *real constant* type in Numerica [Van Hentenryck et al. 1997] or the *bounded real* type in ECL$^i$PS$^e$ [Cheadle et al. 2003]. The solution typically derived by NCSP solvers consists of intervals for the decision variables that contain solution points valid for all values of the data within the input intervals. Semantically, this is a *fully robust* solution. Specific methods to model data uncertainty with NCSPs have been developed in scheduling (e.g. [Narin'yani et al. 2000]). The thrust of the work on NCSPs, however, has been on efficient integration of interval computation methods in CP [Benhamou 1995].

The few applications that use NCSP to model uncertainty (e.g. [Christie et al.

2002]) do so as quantified CSPs (see below). This view is taken because the semantics of a NCSP is not expressive enough to describe and handle uncertainty; quantification is necessary to specify the nature of solution desired. For example, the *definition* of a NCSP does not specify whether the solution should hold for at least one, or for all, values of the interval coefficients. When we add semantics to the NCSP model, in order to specify the desired solution, then the intervals for uncertain real values in the syntax of a NCSP model describe a convex model of uncertainty. Thus understood, the certainty closure framework can exploit techniques from NCSP in the continuous case.

In a *quantified CSP* (QCSP), variables may be arbitrarily existentially ($\exists$) and universally ($\forall$) quantified. As one consequence, a QCSP provides an expressive way of describing uncertainty. The quantifiers allow the precise description of the solution sought (e.g. a robust control policy that holds for all values of a parameter [Zhou et al. 1996]). They allow also the natural expression of bounded uncertainty. For example, existential quantification when some parameter can be chosen by the user, and universal quantification when the exact value of a parameter is unknown or arises in (infinitely) many similar versions [Halpern 2003]. Continuous QCSPs have been applied to model uncertainty due to incomplete or erroneous real-valued data. The *uncertain CSP* model we will define is an instance of the general QCSP formalism. However, the expressive power of QCSPs means it may not always be apparent to the problem modeller how best to represent the uncertainty. Further, the complexity of solving the general QCSP problem is an obstacle. Only in restricted constraint classes are there polynomial time algorithms [Benhamou and Goualard 2000; Ratschan 2006].

Although there is promise in NCSPs and QCSPs for our objective of reliable constraint reasoning, the ideas have yet to be brought together. We find that techniques from other fields, outside CP, are suitable when seeking to achieve this synthesis. Whilst our aim is to tackle ill-defined problems using the CP formalism, researchers in operational research and in robust control in engineering have addressed similar problems. Indeed, the need for non-probabilistic models to address feasibility of and optimality in problems with erroneous and incomplete data has been recognised also in these fields.

> "We hope that future research will also address [in addition to incorporating risk] the issue of how to incorporate ... 'fuzzy' data, i.e., data for which even the mean value is not known and for which one only has range estimates of its value." [Hoffman 2000]

*Robust optimisation* (RO) [Ben-Tal and Nemirovski 1999; El Ghaoui et al. 1998; Hoffman 2000] treats uncertainty in the data as a deterministic, bounded, but unknown quantity, in order to address data uncertainties from the perspective of computational tractability. The outcomes sought from these closure-based models include a robust solution, or a solution that trades-off optimality and reliability, e.g. the best objective value (in the worst case) of all the fully robust solutions. Sharing a similar spirit with RO is *robust control* [Zhou et al. 1996], a field of engineering that aims to operate systems in a way insensitive to faults and measurement imprecision.

In both fields, the uncertain data is described by parameters, whose possible values are captured completely by an *uncertainty set*. For the purpose of tractabil-ity of the resulting model, these uncertainty sets are often composed by means of *convex modelling*, a well established technique whereby uncertain quantities are enclosed with what is known about them: for example, a measurement is enclosed by an interval. Convex modelling has shown its adequacy to lead to reliable solutions [Chinneck and Ramadan 2000; Elishakoff 1995]; indeed, the field is known as *reliable computation*. The most important property of convex modelling is the guarantee that the output contains the true solution or solutions, if any exist. That is, whatever the true value of the data (among the possible realisations encompassed by the model) if a solution exists then it lies within this *complete solution set*.

The *closure paradigm* thus found in reliable computation is potentially valuable in solving models with incomplete and erroneous data. Reliable computation (of which interval analysis over the reals is a concrete instance) is primarily interested in the complete set of solutions. This outcome suits ill-defined problems where the data results from inaccurate or rounded measurements, such as applications in engineering and networking (e.g. [Dovier et al. 2005; Jaulin 2006; Ben-Ameur and Kerivin 2005]). As our case study will show, it gives also valuable insight when diagnosing the reliability of other approaches (see also [Jaulin 2006]).

In summary, the absence of a general framework with unified model and generic resolution forms is problematic from a programming point of view. The user needs to know how to model the problem, usually by choosing a CP versus an OR approach, and which solving method to use accordingly. Our unifying framework aims ultimately at providing a generic modelling formalism, non-probabilistic and tractable, that extends CP and allows one to formulate the ill-defined problem independently of the methods used. In addition to the model of the problem, the framework encapsulates the various types of solution that can be sought, including most robust and fully robust solutions and the complete solution set.

## 3. PRELIMINARY NOTIONS AND BASIC DEFINITIONS

Let $\mathcal{C} = \{c_i\}$ be a finite set of constraints. We represent a solution or set of solutions to a CSP $\langle \mathcal{V}, \mathcal{D}, \mathcal{C} \rangle$ by a conjunction of its constraints, e.g. unary equality constraints. Uncertainty in the data is materialised by uncertainty in the constraint coefficients. A coefficient in a constraint may be *certain* (its value is known) or *uncertain* (value not known). In a classical CSP, of course, all the coefficients are certain. We call an uncertain coefficient a *parameter*. A constraint with parameters is an *uncertain constraint*. Under the closure paradigm, we suppose that the uncertainty in the parameters can be enclosed with certainty, e.g. a set of possible values, an interval. We call the set of possible values of a parameter $\lambda_i$ its *uncertainty set*, denoted $U_i$. We say that a *realisation* of the data is a fixing of all the parameters to values; in related literature, the terms *scenario*, *possible world*, and *context space* can also be found.

The notation $\widehat{\phantom{c}}$ will denote certainty. For an uncertain CSP $P$ (formally defined below), we say that any certain CSP $\widehat{P}$, corresponding to a realisation of the parameters of $P$, is a *realised CSP*, and write $\widehat{P} \in P$. Each uncertain constraint is made certain by a realisation; thus $\widehat{P} = \langle \mathcal{V}, \mathcal{D}, \widehat{\mathcal{C}} \rangle$, where $\widehat{\mathcal{C}} \in \mathcal{C}$ denotes a set of *realised* constraints. In the same way, a realisation of a constraint $c$ will be denoted $\hat{c} \in c$. It is worth noting that an uncertain constraint can have many realisations, as many as the size of the Cartesian product of uncertainty sets involved. Thus, the space of possible realisations is our logical representation of uncertainty.

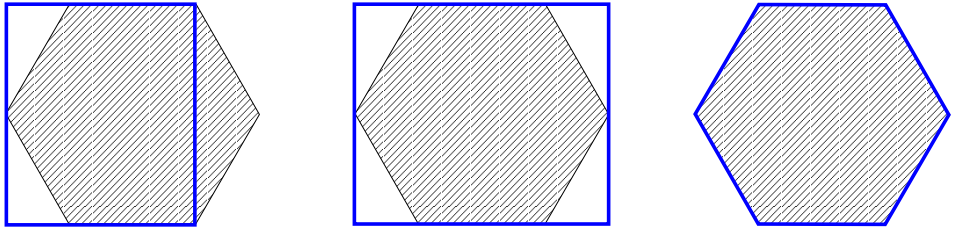

Fig. 1. For a solution set (the shaded polygon), three posited closures: respectively incorrect, correct but not tight, and correct and tight.

Finally, if $r$ is a realisation and $\widehat{P}$ is a corresponding realised CSP, for a solution $s$ of $\widehat{P}$, we say that the realisation $r$ *supports* $s$, and $s$ *covers* $r$. These records of contributing realisations we call *support information* We say that a realisation is *feasible* if the constraints of the problem permit it to ever occur (otherwise *infeasible*). A feasible realisation $r$ is *good* if some solution $s$ exists for the decision variables, given that the parameters have taken their values under $r$ (otherwise *bad*); then of course $s$ covers $r$.

### 3.1 Uncertain Constraint Satisfaction Problem

An *uncertain CSP* extends a classical CSP with an explicit description of the data:

*Definition* 3.1. An *uncertain CSP* (UCSP) $\langle \mathcal{V}, \mathcal{D}, \Lambda, \mathcal{U}, \mathcal{C} \rangle$ is a classical CSP $\langle \mathcal{V}, \mathcal{D}, \mathcal{C} \rangle$ in which some of the constraints may be uncertain. The finite set of parameters is denoted by $\Lambda$, and the set of corresponding uncertainty sets by $\mathcal{U}$.

This definition resembles that of a mixed CSP [Fargier et al. 1994], extending it to the case of continuous data and variables. Moreover, we will define a solution of a UCSP appropriately for reasoning with incomplete and erroneous data.

An uncertain CSP model encompasses arbitrary data types (discrete and continuous) and does not impose a particular representation of the uncertainty set $\mathcal{U}$. How the uncertainty should be represented — a set of values, an interval, an ellipsoid, or otherwise — is a question to be adapted for each computation domain and problem instance. For discussion of the issue of uncertainty set construction, which is outside the scope of this paper, we refer to [Hoffman 2000; Bertsimas and Brown 2006].

*Example* 3.1. *Let $X$ and $Y$ be variables with integer domains $D_X = D_Y = [1, 5]$. Let $\lambda_1$ and $\lambda_2$ be parameters with uncertainty sets $U_1 = \{2, 3, 4\}$ and $U_2 = \{2\}$. Let $c_1$ be the constraint: $X > \lambda_1$, $c_2$ be the constraint: $|X - Y| = \lambda_2$, and $c_3$ be the constraint: $Y - \lambda_1 \neq 1$. Writing $\mathcal{V} = \{X, Y\}$, $\mathcal{D} = \{D_X, D_Y\}$, $\Lambda = \{\lambda_1, \lambda_2\}$, $\mathcal{U} = \{U_1, U_2\}$, and $\mathcal{C} = \{c_1, c_2, c_3\}$, then $\langle \mathcal{V}, \mathcal{D}, \Lambda, \mathcal{U}, \mathcal{C} \rangle$ is a UCSP. Note that, since $\lambda_2$ has just one value in its uncertainty set, it is in fact a constant coefficient. Hence, $c_2$ is a certain constraint; $c_1$ and $c_3$ are both uncertain constraints.* □

### 3.2 Closures: The Solution to a UCSP

In concrete terms, the desired outcome to an ill-defined problem depends on two, linked issues: the requirements of the user and the nature of the uncertainty. As we saw in Section 2, a robust solution, one that holds for as many realisations as possible, is common in CP and OR. However, when the data is uncertain due to inaccurate measurements, for example, we do not want a solution that holds in as many erroneous cases as possible. As another example, in a diagnosis problem, the user simply might want to know whether there exist any realisations at all with solutions (i.e. any good realisations).

A *potential solution* is a complete, consistent assignment to the decision variables that holds for *at least one* realisation of the data. As such, a potential solution is one possibility for a solution to the true problem as modelled by a UCSP. Formally:

*Definition* 3.2. A *potential solution* to a UCSP $P$ is an instantiation to the variables such that there exists a realisation of the parameters, such that all the constraints of $P$ hold. Thus a potential solution is a tuple $\vec{v}$ s.t. $\exists \vec{\lambda_v} : \mathcal{C}(\vec{v}; \vec{\lambda_v})$, where $\mathcal{C}(\vec{v}; \vec{\lambda_v})$ denotes that the conjunction of constraints $\mathcal{C} = \bigwedge_i c_i$ is true under the assignments $\mathcal{V} = \vec{v}$ and $\Lambda = \vec{\lambda_v}$. □

*Definition* 3.3. The *complete solution set* of a UCSP $P$ is the set of all potential solutions to $P$. By abuse of language we sometimes omit 'potential' and just write *solution*. □

Example 3.2. *In Example 3.1, one potential solution is $X = 3 \wedge Y = 1$, because this is a solution to the realised CSP corresponding to $\lambda_1 = 2$.* □

*Definition* 3.4. The *full closure* Cl($P$) of a UCSP $P$ is the set of all solutions such that each is supported by at least one realisation, i.e. the complete solution set. A *closure* in general is a subset of the complete solution set, i.e. $S \subseteq \text{Cl}(P)$. □

Hence, for a UCSP, the notion of CSP solution is replaced by the notion of a closure. We list various types of closures in Section 5.1. Being a set of solutions, any closure can be described by constraints.

Example 3.3. *Let $P$ be the UCSP of Example 3.1, in which the variables $X, Y \in \mathbb{Z}$ have domains $[1, 5]$. The full closure* Cl($P$) *is $(X, Y) \in \{(3, 1), (3, 5), (4, 2), (5, 3)\}$. We can also write it as the disjunction of $(X = 3 \wedge (Y = 1 \vee Y = 5))$, $(X = 4 \wedge Y = 2)$, and $(X = 5 \wedge Y = 3)$.* □

In terms of set inclusion, if one closure encloses another, i.e. $S \subseteq S'$, then $S'$ is a *correct* approximation of $S$: it is correct because no potential solution is omitted. If further $S' = S$, then we say that $S'$ is *tight* to $S$. Figure 1 illustrates.[1]

With the basic notions of the certainty closure framework defined, we next illustrate the principles and value of the framework on the Network Traffic Analysis Problem introduced in Section 1.

---

[1] In the terminology of numerical CSP solving, the concept of correctness is called 'completeness'. The dual property, 'soundness', which means only solutions are included, corresponds to our notion of tightness. We do not adopt this terminology because 'complete' and 'sound' have different meanings in classical CSP solving; and still different meanings in other fields.

## 4. CASE STUDY: NETWORK TRAFFIC INFERENCE

In this section we present an uncertain CSP approach for the *Network Traffic Analysis Problem* (NTAP), and compare it against a prior data correction approach. Given a network with incomplete and inaccurate traffic volume measurements at routers, the NTAP is to determine the minimum and maximum values for each end-to-end traffic flow [Gervet and Rodošek 2000; Rodošek and Richards 2003]. Knowing and managing such traffic flows is a key component to managing many networking components, enabling for instance network performance optimisation and differentiated services [Grossglauser and Rexford 2004]. Various techniques have been proposed for flow profiling but are known to be unreliable unless combined with alternative tools to define, parse, and analyse the flows [Estan and Varghese 2003]. These are time consuming and require domain expertise.

Consider the fragment of a network shown in Figure 2. Four nodes, corresponding to routers and designated `A`–`D`, are shown, together with the aggregated traffic volume on each link. Links are represented by edges in the graph. Each router makes decisions on how to direct traffic it receives, based on a routing algorithm (e.g. OSPF) and local flow information. The routing determines that the traffic flows bidirectionally on each link, apart from `A→C`. This network is taken from [Medina et al. 2002]; we call it `sigcomm4` and use it as a running example.

The traffic volume data is collected by reading router tables at each node over a given time interval (e.g. 20 minutes). As a result, the data information obtained is erroneous. On the link `A→C`, for example, the aggregated traffic volume might measure 565 at $A$ and as 637 at $C$, whereas the true value, equal at both nodes, is presumably somewhere in between. A common approach therefore is to use the median value. Despite the erroneous nature of the problem as received by the network operator, it is known that the true problem must be satisfiable, because the network exists and is executing.

Our objectives are threefold: (1) to simulate the network behaviour in a reliable manner, i.e. to build a model handling the data uncertainty such that a relevant problem is solved; (2) to seek correct bounds tight enough to satisfy the user, i.e. to enclose the true bounds with an outer approximation to the user's satisfaction; and (3) to achieve these first two objectives in a way that is effective in practice.

The initial approach to the problem [Gervet and Rodošek 2000] formulated it as a numerical CSP. Using real-world data collected at routers (no approximation or correction), this model was unsatisfiable. A second model was derived by selecting representative values for the parameters using data correction methods to pre-process the CSP, so ensuring satisfiability of the model. Because the model is linear, a hybrid constraint and linear programming (LP) method was used to find the bounds on the end-to-end traffic flows.

This approach is efficient because it solves a deterministic, well-structured model, but it amalgamates the issues of constraint satisfiability and data errors. Further, there is no indication that the resulting satisfiable model and its solution are reliable. We introduce an uncertain CSP model to investigate the integration of reliable computation concepts in the CP paradigm. It will also enable us to diagnose the actual insights gained by the initial approach, and to provide an alternative reliable approach that remains effective in the presence of uncertainty. For background on the general *traffic inference problem* and data issues in networking, we refer to [Grossglauser and Rexford 2004; Estan and Varghese 2003; Medina et al. 2002].

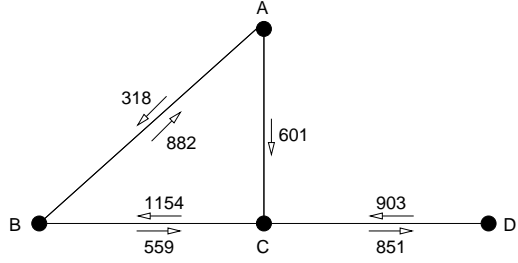

Fig. 2.   Topology and traffic volumes in `sigcomm4`.

## 4.1   Data Correction Approach to the NTAP

4.1.1   *Model.* A network such as `sigcomm4` can be modelled as a numerical CSP as follows. The decision variables correspond to the traffic flow between designated end-point routers. In Figure 2, all four routers are end-points; thus an example of an end-to-end flow is the flow $F_{ac}$ between $A$ and $C$. The variables (decision variables in bold) and constants are:

—**Traffic flow variables.** $\mathbf{F_{ij}} \in \mathbb{R}^+$ is the volume of traffic flowing from node $i$ to node $j$. There are flow variables for each pair $(i, j)$ of routers. Only those for the end-point routers are decision variables.

—**Traffic volumes.** $v_k \in \mathbb{R}^+$ is the measured volume of traffic on edge $k$. This is the main data in the problem, assumed constant and known. In Figure 2, the weights on the edges are the measured aggregated traffic volumes on the links.

—**External traffic volumes.** $T_i^{\text{in}} \in \mathbb{R}^+$ and $T_i^{\text{out}} \in \mathbb{R}^+$ are the measured volume of traffic respectively entering and leaving the network at end-point node $i$.

—**Traffic routing.** $A_{ijk} \in [0, 1] \subset \mathbb{R}$ is the proportion of the flow $F_{ij}$ that uses link $k$. This data is assumed a priori known from the routing.

The constraints in the network problem form a linear flow model. They state conservation of flow at each router (the volume of traffic through each link in each direction is the sum of the traffic entering the link in that direction), plus side constraints. The constraints are of four types:

—**Link traffic constraints.** There is one link traffic constraint on each edge $k$, stating the sum of traffic using the link is equal to the measured volume:

$$\sum_{ij \in k} A_{ijk} \mathbf{F_{ij}} = v_k \qquad \forall k \qquad (1)$$

where the left-hand summation is over all flows $\mathbf{F_{ij}}$ that use edge $k$, i.e. flows i→j with $A_{ijk} = 0$ may be omitted.

—**Traffic conservation constraints.** There are two traffic conservation constraints for each end-point router, stating that the traffic entering the network must equal the traffic originating at the router, and the traffic leaving must equal the traffic whose destination is the router.

$$\sum_j \mathbf{F_{ij}} = T_i^{\mathrm{in}} \qquad\qquad \forall i$$

$$\sum_i \mathbf{F_{ij}} = T_j^{\mathrm{out}} \qquad\qquad \forall j$$

(2)

—**Flow conservation constraints.** In the basic form of the NTAP, there is one flow conservation constraint at each node $j$, of the form:

$$\sum_{k \in e^+(j)} v_k - \sum_{k \in e^-(j)} v_k = T_j^{\mathrm{in}} - T_j^{\mathrm{out}} \qquad\qquad \forall j$$

(3)

where $e^+(j)$ denotes the incoming edges to node $j$, and $e^-(j)$ the outgoing edges. These constraints are redundant due to (1) and (2).

We omit discussion of the side constraints, since, although significant in general, they do not arise for the `sigcomm4` instance of NTAP. The objective functions in the NTAP are to find the min and max of $\mathbf{F_{ij}}$ for each $i \neq j$.

EXAMPLE 4.1. *The NCSP for* `sigcomm4` *has link traffic constraints:*

$$
\begin{array}{ll}
A{\to}B & F_{AB} = v_{AB} \\
B{\to}A & F_{BA} + F_{CA} + F_{DA} = v_{BA} \\
A{\to}C & F_{AC} + F_{AD} = v_{AC} \\
B{\to}C & F_{BC} + F_{BD} = v_{BC} \\
C{\to}B & F_{CA} + F_{DA} + F_{CB} + F_{DB} = v_{CB} \\
C{\to}D & F_{CD} + F_{AD} + F_{BD} = v_{CD} \\
D{\to}C & F_{DC} + F_{DA} + F_{DB} = v_{DC}
\end{array}
$$

(4)

*where we have written the $v_k$ as $v_{ij}$ to explicitly name which routers* `i`{$\to$}`j` *the link $k$ is between; and traffic conservation constraints:*

$$
\begin{array}{ll}
A \text{ origin} & F_{AD} + F_{AC} + F_{AB} = T_A^{in} \\
A \text{ dest.} & F_{DA} + F_{CA} + F_{BA} = T_A^{out} \\
B \text{ origin} & F_{BD} + F_{BC} + F_{BA} = T_B^{in} \\
B \text{ dest.} & F_{DB} + F_{CB} + F_{AB} = T_B^{out} \\
C \text{ origin} & F_{CB} + F_{CA} + F_{CD} = T_C^{in} \\
C \text{ dest.} & F_{BC} + F_{AC} + F_{DC} = T_C^{out} \\
D \text{ origin} & F_{DB} + F_{DA} + F_{DC} = T_D^{in} \\
D \text{ dest.} & F_{BD} + F_{AD} + F_{CD} = T_D^{out}
\end{array}
$$

(5)

*For simplicity, we use none of the redundant flow conservation constraints.* □

If the data ($v_k, T_i^{\mathrm{in}}$ and $T_i^{\mathrm{out}}$) is measured correctly, as in the second column of Table I below, then the NCSP of Example 4.1 is consistent. In fact, there are many solutions for the flows $F_{ij}$ satisfying the constraints since the NTAP is under-constrained in its basic form [Goldschmidt 2000].[2]

4.1.2   *Data/Error Correction and Problem Solving.*   The above model was used in the initial approach to the NTAP. It incorporates directly the measured traffic volumes. Since the measured data led to an inconsistent model, a data correction procedure was used (minimising deviation on the link volumes) in order to reach a satisfiable, deterministic model.

The data correction is performed using a Gaussian error model. It operates in two stages to obtain a single value designed to approximate the true value. Each item of data is considered to be perturbed by two error terms, *gross* and *minor* error. The gross error models entirely fallacious measurements, such as when a router reports 0 but the traffic is non-zero. The minor error models smaller perturbations of the true value by a stochastic $\pm N(0, \sigma^2)$ term. In the data correction, first the global gross error is minimised, then the global minor error.

Empirical studies show that the gross error can be effectively eliminated, provided it occurs rarely [Simonis and Hansen 2002; Feldmann and Rexford 2001; Zhang et al. 2003]. In contrast, minor errors, which tend to be endemic across the data, are nearly impossible to correct systematically.

The corrected model was solved using linear programming. The objective of the NTAP was for bounds for the traffic flows. The bounds, if correctly enclosing the true flows, would indicate the network bottlenecks (close bounds) or sparsity of traffic through the different links (broad bounds). LP was used to derive two bounds per flow: for each flow variable $F_{ij}$, two LPs were solved, with objectives respectively min $F_{ij}$ and max $F_{ij}$.

4.1.3   *Simulating Data Errors.*   To be able to run experimental studies to assess reliability, we need to be able to simulate the NTAP data uncertainty in a realistic and generic manner. Table I gives one data set for the NTAP. In the table, the 'true values' for the data (second column) were simulated exactly from artificially generated demands. The demands are given by [Medina et al. 2002]. From the demands, we calculate simulated traffic flows. From the flows, we calculate the traffic volume measurements. The 'measured' values (third column) are these 'true values' perturbed according to a Gaussian minor error term $\pm N(0, 10)$. For example, on the link A→B, suppose the volume measures as 312 at A and 324 at B. Thus we have two measurements, say $312 \pm N(0, \sigma^2)$ and $324 \pm N(0, \sigma^2)$, of the true value.

The measured values are the input to the NTAP; they correspond to the genuine problem perceived by the user. Despite the true values for the flows being unknown to the user, of course, we can use them in our simulations to evaluate approaches to the problem by comparing the output of the NTAP against them.

EXAMPLE 4.2.   *Consider the* `sigcomm4` *problem. We perturb the true data to the values in the third column of Table I, to simulate 'measured' values. The resulting NCSP is inconsistent; there are no solutions for the flows. We apply error correction to the data in Table I. There are no gross errors, so we use a linear*

---

[2]The essential reason is that the number of flows is quadratic in the size of the network, but the number of constraints (and the number of items of data) is linear.

| Quantity | True Value | Measured | Corrected | Bounded |
|---|---|---|---|---|
| $v_{AB}$ | 318 | 319.67 | 319.9 | [309.98, 327.82] |
| $v_{BA}$ | 882 | 886.20 | 600.97 | [876.39, 894.35] |
| $v_{AC}$ | 601 | 600.08 | 886.2 | [591.93, 612.34] |
| $v_{BC}$ | 559 | 551.45 | 555.51 | [543.30, 562.61] |
| $v_{CB}$ | 1154 | 1151.42 | 1151.42 | [1143.27, 1161.06] |
| $v_{CD}$ | 903 | 904.26 | 904.01 | [896.11, 913.98] |
| $v_{DC}$ | 851 | 853.20 | 853.2 | [842.09, 861.35] |
| $T_A^{\text{in}}$ | 919 | 920.87 | 886.2 | [912.72, 929.02] |
| $T_A^{\text{out}}$ | 882 | 882.85 | 883.32 | [874.70, 891.00] |
| $T_B^{\text{in}}$ | 853 | 853.71 | 870.68 | [845.56, 861.86] |
| $T_B^{\text{out}}$ | 884 | 892.64 | 904.01 | [884.49, 900.79] |
| $T_C^{\text{in}}$ | 918 | 916.43 | 920.87 | [908.28, 924.58] |
| $T_C^{\text{out}}$ | 872 | 870.68 | 853.71 | [862.53, 878.83] |
| $T_D^{\text{in}}$ | 851 | 850.85 | 916.43 | [842.70, 859.00] |
| $T_D^{\text{out}}$ | 903 | 899.24 | 853.2 | [891.09, 907.39] |

Table I. Data in the `sigcomm4` instance of the NTAP. The 'true values' are the traffic volumes generated by the simulated demands. The 'measured' values are those received as input to the NTAP: perturbed from the true values. The 'corrected' values are the results of the data correction procedure on the measured volume values; they should be contrasted with the 'bounded' values obtained by enclosing the uncertainty of the measured values.

*correction term of the sum of the minor errors. The 'corrected value' column shows the result. Observe that in general the correction does not yield values close to the true traffic volumes in the second column. In some cases, such as $v_{BA}$, it 'corrects' in the wrong direction, i.e. away from the true value. Nonetheless, after the data correction, the 'corrected values' make the NCSP of Example 4.1 consistent.* □

### 4.2 Certainty Closure Approach to the NTAP

We now explain our uncertain CSP model of the Network Traffic Analysis Problem and specify the outcome we seek as a solution. The initial data correction model makes two approximations. First, for each parameter it takes one value derived by the error correction. Second, the model assumes simplified routing by equal splitting of traffic, in order, again, to allow its modelling by a single value. In contrast, we seek to simulate the actual network behaviour more closely. We adopt essentially the same constraint model in order to focus on the issues raised by the data uncertainty.

The initial model, being built from the corrected data, is deterministic and tractable, but it does not guarantee to derive correct flow bounds. Moreover, the extent of its unreliable nature and the degree to which the solution can misinform the user are unknown. The key modelling and operational decisions we wanted to investigate were: by bringing reliable computation models in CP, how does it alter a classical CSP model and solution sought? what does it mean in terms of tractability of the constraint model and efficient solving?

4.2.1 *Enclosing the Parameter Uncertainty.* Given the absence of data trends, we seek a non-probabilistic model. We adopt an interval representation for uncertainty sets to enclose the uncertainty in the parameter values. The task of enclosing the uncertainty is an important step in the modelling approach of the closure paradigm. It assumes that the modeller can enclose to her satisfaction the data uncertainty with definite knowledge, using past data and the user's knowledge of the problem [Bertsimas and Brown 2006]. For the NTAP, we proceed as follows.

Since gross errors can be effectively eliminated by automated analysis, we focus on minor errors. Suppose we take two measurements $X$ and $Y$ (at two different routers) for the traffic volume on a link, $X \leq Y$ (without loss of generality), and suppose we adopt an error model $N(0, \sigma^2)$ for the minor errors. Take bounds that encompass the probability mass of the error to a given confidence, defined as 99.5%. This yields intervals for the measurements, $[\underline{X}, \overline{X}]$ and $[\underline{Y}, \overline{Y}]$. Then let $[\underline{X}, \overline{Y}]$ be the bounds we use for the traffic volume. As shown in Table I, all true values but one ($T_B^{\text{out}}$ is slightly below the lower bound) lie inside the bounds. The confidence is considered reasonable by the end-user; a higher confidence could be reached with more emphasis on data analysis (beyond the scope of this paper).[3]

4.2.2   *Constraints.* In the uncertain CSP model of the NTAP the decision variables are the $\mathbf{F_{ij}}$ as before; their domains are the non-negative reals. The data, however, is modelled by parameters rather than constants; there are three types:

—**Traffic volume parameters** $v_k$**.** The uncertainty sets are the bounds given by the above convex closure procedure.

—**External traffic volume parameters** $T_i^{\mathbf{in}}$ and $T_i^{\mathbf{out}}$**:** similar to the traffic volumes.

—**Traffic splitting parameters** $A_{ijk}$**.** For flows with a single path (and for links on a flow with multiple paths where there is only a single choice), the parameters are constants: the uncertainty sets are singleton values 0 or 1. For flows where splitting occurs between two paths, the uncertainty sets are the intervals $[0.3, 0.7]$; in general, for $r$ paths, the uncertainty sets are $[\frac{0.6}{r}, \frac{1.4}{r}]$. Although there is little empirical knowledge of the exact distribution, it is considered safe to suppose that, when not split equally, traffic splits between 30% and 70% [Simonis 2003][4].

Thus, the constraints are (1)–(3), as in the error correction model, but $v_k$, $T_i^{\text{in}}$ and $T_i^{\text{out}}$, and $A_{ijk}$ are now parameters. Observe that some constraints, such as (1) are now *non-linear* (quadratic, due to the products of variables and parameters).

4.2.3   *Full Closure.* The outcome sought from the UCSP model, like in the data correction model, describes bounds for the traffic flow values. We obtain an interval for each variable that describes an approximation of the complete solution set, i.e. an approximation of the full closure. We call such an outcome to the UCSP model the *projected full closure* (onto the domains of the decision variables). If the (projected) full closure is non-empty, we can advise the user that there is a solution to the problem corresponding to at least one possible realisation of the data. If, on the other hand, the full closure contains empty intervals for some variables, we know that the constraint model itself is inconsistent.

---

[3]Note that a 100% confidence level would lead to infinite bounds.
[4]The distribution of split traffic depends on a number of factors, including the duration of traffic sampling, the configuration of the routers, and the routing protocol itself.

4.2.4  *Model Approximation.*  The quadratic nature of the uncertain constraints in the UCSP model means that we can no longer apply LP directly to derive the sought bounds. Given the complexity of the problem, while investigating methods to its solving we considered an approximation to the model that retains correctness while reducing the computational complexity: namely, independence of the parameters. In the NTAP, the parameters are independent except for those that model the traffic routing (i.e. the [0.3, 0.7]): if 40% of traffic goes one way at a two-way junction, 60% must go the other way, by flow conservation. Besides independence, the initial approach assumed equal splitting of traffic, which is certainly a false assumption. As we see later, for the NTAP assuming independence (only) proved pragmatic and reasonable. It results in less tight but still correct intervals for the traffic flows. Tractably addressing parameter dependency is a challenging research topic in all forms of data uncertainty and is part of our future work.

### 4.3  Solving the UCSP for the Projected Full Closure

We investigated a number of different approaches to solve the UCSP, determining how they differ in their computational complexity and tightness of the solution bounds. These methods include directly applying NCSP and QCSP algorithms, and applying reliable computation techniques. Since the uncertain constraints are well-structured and the closure does not require support information, the latter proved to be the most adequate to derive the projected full closure effectively. It is presented first. This approach seeks a transformation of the UCSP model to an equivalent certain CSP, and solves the latter using LP.

The main insight is that the UCSP is an instance of an *interval linear system*:

*Definition* 4.1. Let $\mathcal{V}$ be a set of $n$ variables over $\mathbb{R}$, and $\mathcal{C}$ be a set of $m$ linear constraints (equalities or inequalities) for $\mathcal{V}$. Writing each constraint in normal form, let $A \in \mathbb{R}^{m \times n}$ be the matrix of left-hand sides, and $b \in \mathbb{R}^m$ be the vector of right-hand sides. Let $R$ be the list of $m$ relations, one for each constraint; $R_i \in \{<, \leq, =, \geq, >\}$, $i = 1, \ldots, m$. Then an *interval linear system* (ILS) [Neumaier 1990] induced by $\mathcal{C}$ on $\mathcal{V}$ is a tuple $\langle \mathbf{A}, R, \mathbf{b} \rangle$, where $\mathbf{A} \in \mathbb{IR}^{m \times n}$ is an interval matrix $[\underline{A}, \overline{A}]$ with $A \in \mathbf{A}$, and $\mathbf{b} \in \mathbb{IR}^m$ is an interval vector $[\underline{b}, \overline{b}]$ with $b \in \mathbf{b}$.  □

EXAMPLE 4.3. *This system of constraints* $\mathbf{A} \cdot V \, R \, \mathbf{b}$ *describes an ILS:*

$$\mathbf{A} = \begin{pmatrix} [-2, 2] & [1, 2] \\ [-2, -1] & -1 \\ 6 & [1.5, 3] \end{pmatrix}, V = \begin{pmatrix} V_1 \\ V_2 \end{pmatrix}, R = \begin{pmatrix} \leq \\ = \\ = \end{pmatrix}, \text{ and } \mathbf{b} = \begin{pmatrix} [3, 4] \\ [-5, 5] \\ [4, 15] \end{pmatrix}$$

The key observation that enabled us to solve our UCSP model effectively, unlike the general case, is that the variables are all non-negative; hence their values all lie in the positive orthant.[5] This special case of an ILS is called a *POLI* system:

*Definition* 4.2. A *positive orthant interval linear system* is an interval linear system in which the natural domain of the variables are non-negative. Thus, the solution set lies within the positive orthant of $\mathbb{R}^n$.  □

---

[5]The positive orthant is the intersection of the non-negative halfspaces. In 2D, for instance, the positive orthant is the upper-right quadrant.

Several robust optimisation approaches exist for interval linear systems. Two are the robust counterpart (RC) methodology [Ben-Tal and Nemirovski 1999] and interval linear programming (ILP) [Chinneck and Ramadan 2000]. In both, the solution sought is a bound on the objective function. In RC, the solution describes the best worst-case objective that satisfies all realisations of the constraints. In ILP, best and worst optima values for the objective function are found, together with a witness realisation under which they occur.

The NTAP, however, is a pure satisfaction problem, and the solution we seek is different to that of both RC and ILP. Moreover, we have a restricted form of ILS which the solving method should exploit. Thus neither approach is directly suitable. However, we observe that both find their sought solution by transforming the uncertain problem to an auxiliary equivalent problem, and solving the latter.

4.3.1 *Algorithm.* Since the outcome sought for the NTAP is bounds on the traffic flows, it suffices to compute the *interval hull* $\Box\Sigma\,(\mathbf{A},\mathbf{b})$ (the projected full closure for an ILS). There are three steps to the algorithm. First, equalities are rewritten as pairs of inequalities. Second, a transformation is applied that results in an *equivalent CSP*, equivalent in terms of the solution set sought. Third, the interval hull of the equivalent CSP is derived. Note that the intervals throughout and the solution set itself may be unbounded.

**Step 1: Rewrite equalities**. The required form is to have linear constraints with interval uncertainty. Thus, each equality constraint is to be replaced by a pair of inequalities as illustrated below.

EXAMPLE 4.4. *Consider the concrete NTAP link traffic constraint $\mathtt{A}\!\to\!\mathtt{C}$ in (4), with flow splitting parameters. It is transformed from:*

$$F_{AC} + [0.3, 0.7]F_{AD} = [591, 613] \tag{6}$$

*into the pair of constraints*

$$
\begin{aligned}
F_{AC} + [0.3, 0.7]F_{AD} &\geq [591, 613] \;\wedge \\
F_{AC} + [0.3, 0.7]F_{AD} &\leq [591, 613]
\end{aligned}
\tag{7}
$$

*and step 2 will further derive (by Proposition 4.3) the constraints:*

$$F_{AC} + 0.7F_{AD} \geq 591 \;\wedge\; F_{AC} + 0.3F_{AD} \leq 613 \tag{8}$$

**Step 2: Transformation of UCSP to an equivalent CSP**. The transformation replaces the matrix $\boldsymbol{A}$ and vector $\boldsymbol{b}$ by non-interval versions.

PROPOSITION 4.3. *The full closure of a POLI system $L = \langle \boldsymbol{A}, R, \boldsymbol{b} \rangle$ is the complete solution set of the numeric linear inequality system $A'\,x \leq b'$, where $A' \in \mathbb{R}^{m\times n}, x \in \mathbb{R}^n,$ and $b' \in \mathbb{R}^m$ are given by the following.*

$$
((A')_i, b'_i) = \begin{cases} (\underline{\boldsymbol{A}_i}, \overline{\boldsymbol{b}_i}) & \text{if } \{<, \leq\} \in R_i \\ (-\overline{\boldsymbol{A}_i}, -\underline{\boldsymbol{b}_i}) & \text{if } \{>, \geq\} \in R_i \end{cases}
\tag{9}
$$

PROOF. We suppose $\{<, \leq\} \in R_i$. The $\{>, \geq\} \in R_i$ case is similar, and the statement for it follows because $-1 \times (Ax \leq b) = -Ax \geq -b$.

Let $S_1 = \Sigma(\mathbf{A}, \mathbf{b})$ be the complete solution set of $L$, and let $S_2$ be the complete solution set of $A' x \leq b'$, where $A' = \underline{\mathbf{A}}$ and $b' = \overline{\mathbf{b}}$. We show that the two solution sets coincide: i.e. $x \in S_1$ iff $x \in S_2$. Since $L$ is a POLI system, $x_i \geq 0 \; \forall i$.

First, suppose $x \in S_2$. Since $A' \in \mathbf{A}$ and $b' \in \mathbf{b}$, by the definition of $\Sigma(\mathbf{A}, \mathbf{b})$, $x \in S_1$. Second, suppose $x \in S_1$. This means $\exists A \in \mathbf{A}, b \in \mathbf{b}$ such that $Ax \leq b$ for $x \geq 0$. Now each row of this numeric linear inequality system, $A_i x \leq b_i$, is a $n$-ary monotone constraint [Zhang and Yap 2000]. The inequality holds true if the LHS is decreased or if the RHS is increased, or both. Observe that $b_i \leq \overline{\mathbf{b}_i}$ and $A_i \geq \underline{\mathbf{A}_i}$. Moreover, for fixed $x$, since $x \geq 0$, $A_i x \geq \underline{\mathbf{A}_i} x$. Hence $\underline{\mathbf{A}_i} x \leq \overline{\mathbf{b}_i}$, and so $x \in S_2$. Since the same holds for each row of the matrix $A'$, this completes the proof. $\square$

Thus the transformation produces an equivalent model, i.e. whose solution set is the full closure to the POLI system.

EXAMPLE 4.5. *Continuing Example 4.3, the initial ILS is transformed into the equivalent CSP $P'$ with the same complete solution set, with constraints $A' \cdot V \leq b'$,*

$$A' = \begin{pmatrix} -2 & 1 \\ -2 & -1 \\ 1 & 1 \\ -6 & -3 \\ 6 & 1.5 \end{pmatrix} \text{ and } b' = \begin{pmatrix} 4 \\ 5 \\ 5 \\ -4 \\ 15 \end{pmatrix} \tag{10}$$

*The first constraint, for instance, is transformed from $[-2, 2]V_1 + [1, 2]V_2 \leq [3, 4]$ to $-2V_1 + V_2 \leq 4$. The decision variables remain unchanged. Observe the increase in the number of constraints (from three in $P$ to five in $P'$), the result of Step 1.* $\square$

**Step 3: Derive the interval hull.** The full closure of an ILS can be represented precisely as a halfspace description of a polytope. The action of the constraints on the variable domains is equivalent to the projection of the convex hull of the polytope onto each axis [Oettli 1965].

In practice we can use linear programming to find the interval hull directly. We do not calculate the exact form of the full closure (as a convex hull), but go straight to the approximation of the outer hyperbox. This corresponds to the projected full closure. For each variable $V_j$, $j = 1, \ldots, n$, we solve two LPs which differ only in their objectives, $\min V_j$ and $\max V_j$, subject to the constraints given by the equivalent problem. With $2n$ applications of simplex we have tight, reliable bounds. For Example 4.5, this gives $\mathrm{Cl}(P) \subset ([0, 2.5], [0, 4.67])$, shown in Figure 3.

This use of LP to find bounds is the same as the initial data correction approach to the NTAP. The difference is that LP is applied to a NCSP obtained from a reliable UCSP model, rather than from data correction.

### 4.3.2 Correctness

PROPOSITION 4.4. *Transformation and repeated LP computes the tightest certain bounds possible, and does so in worst case time $\mathcal{O}(mn^3)$.*

PROOF. The transformation from the uncertain CSP, or ILS, to the equivalent CSP can be done in $\mathcal{O}(2mn)$ operations, and its correctness follows from Prop. 4.3. The constraints derived are linear and the solution space of their conjunction intersected with the positive orthant is a convex polytope, possibly unbounded [Aberth 1997]. Projection onto each normal axis immediately gives the interval hull. The bounds so obtained are tight to the solution set since the maximum and minimum possible value of each variable are found by the simplex iterations. The projection can be done with $2n$ iterations, at expected cost $\mathcal{O}(3mn^2)$ each [Schrijver 1986]. □

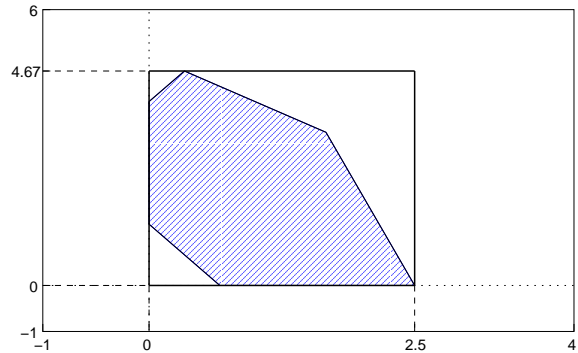

Fig. 3. Tight and certain bounds: the full closure is the shaded region. We compute the tight outer box shown: $[0, 2.5] \times [0, 4.67]$.

Note that the bounds are as tight as possible as far as interval bounds as sought for the NTAP. Were it required, we could seek alternative representations of the full closure, other than its projection on the variable domains — for instance, a union of intervals. The trade-off of the greater fidelity of such representations, i.e. the tightness of the representation of the closure, is a higher computational cost.

4.3.3 *Evaluation of CP Solving Methods.* We briefly describe alternative approaches to derive the projected full closure to the NTAP modelled as a UCSP. Two sources of approaches to such a POLI system are from numerical CSP and quantified CSP.

A numerical, continuous-data CSP attaches real intervals to the variables. Hence, NCSP programming systems that admit the use of interval coefficients can be used to derive the UCSP model; one of these is the `ic` library of ECL$^i$PS$^e$ [Cheadle et al. 2003]. However, the default inference rules in such systems typically derive boxes (i.e. Cartesian products of intervals) that are true for *all* values of the parameters within their intervals. This corresponds to a fully robust solution, whereas we seek the complete solution set, where each point is true for *some* values of the parameters. However, we can benefit from the interval methods embedded within NCSP solvers by modelling the parameters with existentially quantified variables rather than with interval coefficients. To derive a box that encloses the complete solution set requires search (here, domain splitting) as well as interval narrowing. The main drawbacks are the worst case computational cost (exponential in time when performing search), and the loose bounds when search is omitted.

We also investigated solvers for quantified NCSPs. General analytic solvers for quantified NCSPs are too restricted in the classes of constraints they can handle, or too inefficient, to solve sizable POLI systems. One exception is the approximate solver `AQCS` [Ratschan 2000]. Because it gives up the demand for an exact solution, and imposes certain (weak) conditions on the CSP, it is effective enough to be used for ILS problems. It combines propagation over real first-order constraints with search. However, the output of `AQCS` on the NTAP was coarse, even when the accuracy is set to the maximal floating-point representation available. On such a well-structured problem, we concluded that a dedicated solving method is preferable to a general purpose solver.

| Problem | End points | Routers | Links | Links/router |
|---------|-----------|---------|-------|--------------|
| `toy2` | 2 | 5 | 12 | 2.40 |
| `sigcomm4` | 4 | 4 | 7 | 1.75 |
| `dexa` | 24 | 51 | 118 | 2.31 |
| `as3967` | 22 | 79 | 294 | 3.72 |
| `as1755` | 23 | 87 | 322 | 3.70 |
| `as3257` | 49 | 161 | 656 | 4.07 |

Table II. Problem instances of academic and real-world networks.

### 4.4 Implementation and Empirical Evaluation

We assessed the benefits of a UCSP model with uncertainty sets against the pragmatic data correction approach. For both approaches, the modelling and solving used the ECL$^i$PS$^e$ CLP platform. The experiments were conducted on a 2GHz Linux PC with 1GB of memory, using ECL$^i$PS$^e$ version 5.6; timings are in seconds. The linear solver was the primal simplex of Xpress-MP version 14.21.

The UCSP model was implemented using the `ic` interval computation library, which provides a *bounded real* datatype for coefficients: an interval representing an unknown real value. Using `ic`, we can model constraints such as the link traffic constraint A→C in (4) simply by: `F_AC + F_AD = 591__613`. That is, uncertain parameters are written naturally; explicit naming is unnecessary unless desired.

4.4.1 *Methodology.* We studied six benchmark networks, in order to evaluate the insights gained by our approach to the NTAP in terms of solution quality and practicality of the approach. We generated simulated data according to the process described in Section 4.3. The results are averaged over 100 runs.

The benchmarks are characterised as follows:

—**Topology.** `toy2` and `sigcomm4` are small test networks; `sigcomm4` from [Medina et al. 2002] was shown in Figure 2. The other problem instances are real-life networks. `dexa` is the topology of a global ISP.[6] The networks denoted `as` are also operational ISP networks, obtained by using the *rocketfuel* inference tool [Spring et al. 2002].

—**Routing.** Traffic is routed according to a known algorithm[7]; routing information is known, fixed input data to the NTAP.

—**Traffic.** For each network, we generated a set of random demands. From the demand and the routing, traffic volume measurements were calculated for each link. The measurements were then perturbed and some were discarded, resulting in an inconsistent and incomplete data set. This was the input to the NTAP.

---

[6]Due to commercial confidentiality, we cannot report the exact topology of `dexa`.

[7]The details are not important for us here.

| FLOW | NO UNCERTAINTY | | | PERTURBED DATA | | |
|---|---|---|---|---|---|---|
| | true flows | LP bounds | full closure | initial method (single value) | initial method (bounds) | full closure |
| A→B | 318 | [318, 318] | [309, 327] | 319.9 | **[319.90, 319.90]** | [309, 328] |
| A→C | 289 | [22, 581] | [2, 609] | 577.48 | **[21.97, 577.48]** | [1, 608] |
| A→D | 312 | [20, 579] | [0, 591] | 23.49 | [23.49, 579.00] | [4, 591] |
| B→A | 294 | [294, 294] | [277, 311] | 298.20 | **[298.2, 298.20]** | [282, 315] |
| B→C | 292 | [0, 559] | [0, 568] | 0.0 | [0.0, 555.51] | [0, 563] |
| B→D | 267 | [0, 559] | [0, 568] | 555.51 | [0.0, 555.51] | [0, 563] |
| C→A | 305 | [305, 305] | [285, 325] | 305.0 | [305.0, 305.0] | [285, 325] |
| C→B | 289 | [289, 289] | [264, 314] | 286.42 | **[286.42, 286.42]** | [271, 316] |
| C→D | 324 | [324, 324] | [304, 344] | 325.01 | **[325.01, 325.01]** | [305, 340] |
| D→A | 283 | [283, 283] | [274, 292] | 283.0 | [283.0, 283.0] | [274, 292] |
| D→B | 277 | [277, 277] | [268, 286] | 277.0 | [277.0, 277.0] | [268, 286] |
| D→C | 291 | [291, 291] | [266, 316] | 293.20 | **[293.20, 293.20]** | [266, 316] |

Table III. Bounds for the flows in the `sigcomm4` instance of the NTAP. The perturbed data is that in Table I. Almost all single flows obtained by the initial method differ from the true flows. Even if bounds are sought with the initial method, some do not enclose the true flows (these are shown in bold). In contrast, the reliable bounds from the UCSP always enclose the true flows.

—**Errors.** The data errors were created with our error model explained earlier; recall that the minor errors thus have a Gaussian $N(\mu, \sigma^2)$ distribution.

### 4.4.2 *Results.*

*Effect of UCSP model.* We first investigated the initial approach and the UCSP approach with respect to the *reliability* of the model and solution produced. The projected full closure for `sigcomm4`, for the data in Table I, is shown as the final column of Table III. Although calculations are performed on reals, we sometimes show the bounds as integers (rounding bounds outward), for ease of presentation.

The table reports six columns of values or bounds for the flows. The first three columns of data report hypothetical approaches to the NTAP, where we assume there is no data uncertainty: they are based on the true values for the traffic volumes as first introduced in Table I. With uncertainty removed, the results of this part of the experiment evaluate the bounds derived by both the initial and UCSP approaches.

The 'true flows' column are the generated flows from [Medina et al. 2002], representing values for the flows without data uncertainty. The 'LP bounds' column shows the bounds produced by LP, indicating the tightest possible bounds that LP can produce from the deterministic model without uncertainty. The bounds describe a single value in some cases and are loose approximations in others. This reflects upon the under-constrained nature of the problem as specified (recall there are many more end-to-end flows than constraints) [Goldschmidt 2000].

The next column, 'full closure', shows the results of applying the certainty closure to the model without data uncertainty. The difference between the 'LP bounds' and the 'full closure' column, a slight broadening of the bounds, comes from LP solving a UCSP where equalities have been rewritten into inequalities, and further transformed to a linear form. The widening is marginal. The full closure bounds enclose the 'LP bounds'.

| | | Initial Method | Certainty Closure | |
| Variance | Error % | % valid (bounds) | lower bound | upper bound |
|---|---|---|---|---|
| 0 | 0 | 100 | 67.84 | 132.28 |
| 0.1 | 0.03 | 58.33 | 67.24 | 132.96 |
| 0.5 | 0.17 | 58.33 | 67.01 | 133.20 |
| 1 | 0.34 | 58.33 | 66.16 | 134.18 |
| 5 | 1.69 | 58.33 | 62.20 | 140.15 |
| 10 | 3.39 | 58.33 | 61.85 | 140.65 |
| 20 | 6.78 | 58.33 | 61.37 | 141.37 |
| 50 | 16.94 | 58.33 | 60.39 | 142.78 |
| 100 | 33.89 | 58.33 | 59.32 | 144.38 |
| 200 | 67.78 | 53.67 | 57.77 | 146.59 |
| 500 | 169.45 | 26.83 | 54.73 | 150.98 |

Table IV. Error analysis for `sigcomm4`. 'error %' is the magnitude of the mean perturbation relative to the mean flow size.

The last three columns of Table III are based on the actual erroneous input to the NTAP: the perturbed data (the 'measured' traffic volume values in Table I). The column 'single value' shows the single flow values derived from the corrected data (the 'corrected' values in Table I). The reason why flow bounds are sought is seen in the 'single value' column. When a single LP is solved for each flow with the objective to minimise (globally) the error according to the data correction model, the derived solution is wrong. The values do not match the true flow values in most cases (comparing columns 2 and 5): in 75% of cases, in fact. The values correspond to one extremum in the bounds approach because the global objective seeks a satisfiable solution to an under-constrained problem.

The sixth column shows traffic flow bounds, based on the corrected data according to the error correction model. For five flows shown in bold, the true flow value is not contained in the bounds. Further, the equal upper and lower bounds fallaciously suggest a definite value. In contrast, the closure bounds, derived from the 'bounded' values in Table I and shown in the final column, always enclose the true flow values.

*Effect of uncertainty.* Table III shows that the flows derived by the UCSP approach in the presence of perturbed data are reliable and comparable in tightness to those derived without uncertainty. Our second experiment investigated in more detail the effect of inaccuracy in the data on the tightness of the solution bounds. For both models, do the bounds get looser as the uncertainty grows; and if so, how does it impact the meaningfulness of the bounds?

Table IV evaluates variance of flow tightness with the amount of uncertainty. We specified the data uncertainty by the stochastic minor error $\pm N(\mu, \sigma^2)$. The mean $\mu$ is zero; the variance $\sigma$ increases from 0 (no perturbation) to 500. Since the 95% percentiles of the normal distribution lie at $\pm 1.96\sqrt{\sigma}$, the latter rows in the table exhibit substantial minor error, verging on gross error. The second column shows the magnitude of the mean error relative to the mean flow size.

The '% valid' column is for flow *bounds* derived by the data correction method. It is the proportion of bounds that contain the true flow values. With no uncertainty (first row), the bounds always contain the true value, as expected. However, with any uncertainty, the validity of the flow bounds falls to just under 60%. The validity is constant whether there is modest or large amounts of uncertainty. Only when the magnitude of uncertainty becomes very great (more than 50% of the true data in magnitude) does the validity fall further. It seems that data correction can produce a 'corrected' data set that is a fixed error distance from the true data, until the point when it is overwhelmed by the uncertainty. Beyond this point, there is no distinction between minor and gross errors.

| Problem | Lower Bound | Upper Bound | Time |
|---------|-------------|-------------|------|
| `toy2` | 100% | 100% | 0.06 |
| `sigcomm4` | 17% | 184% | 0.16 |
| `sigcomm4'` | 68% | 132% | 0.13 |
| `dexa` | 2.8% | 26062% | 53.4 |
| `as3967` | 0% | 2454% | 224 |

Table V. Results on academic and real-world networks. The bounds are given as the percentage of the true flow values (averaged over all flows); 100% is perfect.

The last two columns give the lower and upper bounds of the certainty closure, as a percentage of the true flow values. They show that the certainty closure produces bounds that widen monotonically as the uncertainty increases. However, they widen much more slowly than the increase in uncertainty, showing that the certainty closure produces solutions relatively insensitive to the *magnitude* of uncertainty.

*Indicators of model constrainedness.* Finally, we evaluated our transformation-based solving approach on different networks characterised by their topology, size and traffic. Table V shows the results for the four problem instances defined in Table II above, and one additional network we describe below. The bounds are shown as the percentage of the values of the true flows (for no data uncertainty). Thus 68% indicates the derived flow value is about 0.7 of the true flow value. As in Table IV, the closer to 100%, the tighter the bounds.

Table V indicates how the UCSP model gives insightful results on real world networks involving more than 50 routers, 300 links, and 20 end-points. CPU times, given in seconds, account for the transformation step and the two LP runs for each decision variable. Experiments showed that the bulk of the time reported is spent in garbage collection, indicating the present implementation is space-bound, not time-bound; a more efficient implementation of our current prototype is expected to substantially reduce the runtime.

Observe that the bounds vary tremendously: tight for `toy2` and much less tight for `dexa` and `as3967`. We infer that (1) the model as specified is under-constrained and thus either more data or network information is required to derive informed bounds, (2) some links are under-used (suggesting potential for redesigning the network topology), or (3) there are no clear bottlenecks (saturated links).

Since an overwhelming amount of information is available from different sources and network devices (e.g. routers), a complex task when dealing with network design is to identify relevant data that will add useful information to the model, without being overwhelmed with an excessive volume of data. This is a research field in its own right [Grossglauser and Rexford 2004].

The data correction model did not consider such additional data. However, we discovered that the bounds can be tightened by improving the constraint model, for example by adding constraints from *LSP counters* [Davie and Rekhter 2000]. LSP counters provide additional information about the traffic flows, making the model less under-constrained, and so leading to tighter bounds. We used these additional LSP constraints for `sigcomm4` to illustrate on our running example, leading to the bounds shown on the third row `sigcomm4'`. The tighter bounds with the LSP constraints lead to a more informative discussion.

### 4.5 Discussion

Our certainty closure approach adopts a reliable model and leads to reliable solutions. We have diagnosed that the model of the initial approach did not capture the genuine problem. In contrast, the certainty closure presents a satisfiable model (provided, of course, that solutions exist to the true problem) that is also a reliable model. Compared to the initial approach to the NTAP, our approach leads to guaranteed quantitative results. Section 5 will formalise and generalise the solving method of our UCSP model into a *transformation resolution form*.

Second, our approach extends a classical CSP by accounting for data uncertainty while at the same time preserving the properties of a deterministic model deprived of uncertainty. It separates data and constraint issues. When the interval found for a flow variable is empty, we can infer that the UCSP model is unsatisfiable, implying that the constraint model is unsatisfiable not due to inaccurate data values. In comparison, data correction is designed to always produce a satisfiable model; it finds a solution whether or not one exists to the true problem.

Third, our approach enables insight into the sources of bottlenecks in the network at no extra computational cost. For instance, tight bounds for a flow indicate that in all possible solutions the traffic behaves in a similar way, while broad bounds indicate a volatile flow. Knowledge of this kind is sought by network operators [Feldmann et al. 2001]. Likewise, a lower bound on a link close to its capacity indicates a critical, saturated link. The bounds arising from the initial approach do not reliably support such inferences. Finally, if *all* the bounds are broad, this can imply that the current data is insufficient, and thus the constraint model is under-constrained, or that the network is under-used.

As an extension of CSP models, inspired by models and techniques from reliable computation, the model and operational behaviour we have presented for the NTAP problem are not defined within the CP framework. Thus in the following section we introduce the remainder of our generic formal framework that defines a UCSP model, and give the operational semantics independently of a specific method but rather in terms of algorithmic properties.

## 5. THE CERTAINTY CLOSURE FRAMEWORK

The certainty closure framework is composed of two main elements: (1) an extension of the CSP formalism to represent ill-defined constraint problems with incomplete or erroneous data, and (2) the solving of such models for a suitable outcome, by means of resolution forms. The NTAP case study presented in the last section illustrated the closure paradigm. The main objective of the framework is to offer a reliable, tractable, and non-probabilistic approach to constraint reasoning under uncertainty. The certainty closure framework achieves this objective by extending the CP formalism with the closure paradigm.

*Complexity.* An uncertain CSP is a restriction of a general quantified CSP if we view parameters as existentially quantified variables. Recall that the decision problem for a general QCSP is PSPACE-complete, and for many constraint classes where a CSP is tractable, a QCSP is not [Boerner et al. 2003]. Fortunately, the quantifiers in a UCSP are not arbitrary but are restricted to an existential pair: the question 'Does there exist a potential solution to a UCSP?' is:

$$\exists \vec{v} \, \exists \vec{\lambda_v} : \mathcal{C}(\vec{v}; \, \vec{\lambda_v}) \tag{11}$$

as we saw in Definition 3.2.

Further, provided we retain the ordering in the sentence (11), logically we can abolish the distinction between parameters and variables. That is, letting $x$ and $y$ be variables, (11) is equivalent to:

$$\exists x \, \exists y : \mathcal{C}(x; y) \iff \exists x, y : \mathcal{C}(x; y) \iff \exists z : \mathcal{C}(z) \tag{12}$$

and, changing variables to $z$, this is the decision problem for a classical CSP. This shows that the decision problem for a UCSP is NP-complete. Taking the idea of treating the parameters as variables, we can find a potential solution by solving the resulting CSP. However, it does not necessarily follow that, operationally, this is the best means to find potential solutions.

A second problem of interest to us is the complexity of finding the complete solution set of a UCSP, i.e. finding all potential solutions. Finding all solutions to a classical CSP does not fit into a complexity class in the traditional sense; its characterisation depends on how the output is returned.

One way to characterise the enumeration problem is as a function problem, i.e. a function that, given the formula as an input, produces the list of solutions. Since the output list is of exponential size (in the input), in general, we can say that the problem is FEXPTIME (functional EXPTIME). For a more precise characterisation, we refer to [Schnoor and Schnoor 2006].

## 5.1  Other Closures

In the NTAP case study we were concerned with the full closure. This outcome fitted the desire of the user for bounds on the possible traffic flows. However, depending on the user requirements, alternative solutions might be sought as the outcome to a UCSP. In our formalism, such solutions are specified by properties on individual elements of a set of potential solutions (i.e. closure), or by properties on the whole closure. In summary we have:

—**Full closure.** The set of all solutions that each cover at least one realisation. Example usage: behaviour guarante across all possible solutions; diagnosis of the reliability of approximation methods.

—**Robust set.** A set of solutions such that each cover *all* realisations (not just at least one). A robust set closure is maximal if the cardinality of this set is maximal among all such sets. Example usage: conformant planning.

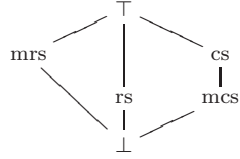

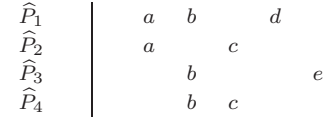

Fig. 4. Simple hierarchy of closures. At the top of the lattice is the full closure, at the bottom the empty closure. Illustrated in the middle are a covering set (cs), a minimal covering set (mcs), a robust set (rs), and the most robust solution (mrs).

Fig. 5. Realised CSPs (denoted $\widehat{P}_1$–$\widehat{P}_4$) and their feasible solutions (denoted $a$–$e$).

—**Most robust solution.** A single solution that covers the maximal number of realisations, of all single solutions. Example usage: robust solution that must be a single solution and not a set of solutions, e.g. schedule for a staff roster [Manandhar et al. 2003].

—**Covering set.** A set of solutions that together cover all realisations. A covering set closure is *minimal* if the cardinality of this set is minimal among all such sets. Example usage: robust solution covering every eventuality, as in contingent planning [Yorke-Smith and Guettier 2003].

To illustrate how they relate in terms of sets of potential solutions, a simple hierarchy of those closures is shown in Figure 4. They form a lattice under set inclusion. The full closure is the top of the lattice, and the empty closure (the empty set) is the bottom.

EXAMPLE 5.1. *Consider the abstract UCSP depicted in Figure 5. There are four realised CSPs, denoted $\widehat{P}_1$–$\widehat{P}_4$, and five potential solutions, denoted $a$–$e$. The top of the lattice hierarchy is the set $\top = \{a, b, c, d, e\}$. The most robust solution is $b$; while uniquely maximal, it does not cover all realisations ($\widehat{P}_2$ is uncovered). Since there is no solution that covers all realisations, the (maximal) robust set is empty. However, there are two covering sets of minimal cardinality, $\{a, b\}$ and $\{b, c\}$.* ☐

These types of closure suit satisfaction problems. When the ill-defined problem includes an objective function, additional types of closures can be defined. For instance, in robust optimisation, the best worst-case solution (a compromise between optimality and reliability) is sought. In this article we focus on satisfaction problems; the extension of the certainty closure framework to uncertain constraint optimisation problems defines new types of closures [Yorke-Smith and Gervet 2005].

## 5.2 Deriving Closures of a UCSP

A UCSP adds expressive power and flexibility to a CSP but, depending on the closure demanded, is harder to solve. We now present means to derive different types of closures of a UCSP. We show how existing suitable algorithms from different fields are valuable within the unifying framework we present. Our aim is for polynomial methods, and, secondarily, for *tight* approximations that preserve the closure (recall Figure 1). Where possible, we would like to exploit existing methods for CSP solving, and leverage concepts and algorithms for general QCSPs.

Even though UCSPs are a restricted form of QCSP, our objective of deriving a closure effectively has lead us to consider dedicated techniques with greater ef-ficiency than general QCSP solving. This does not exclude the use of generic methods where advantageous. For instance, we can use quantified arc consistency [Bordeaux and Monfroy 2002; Mamoulis and Stergiou 2004; Gent et al. 2005] as a preprocessing step.

In the next two sections we define two resolution forms — two possibilities to move from a UCSP to a closure. We say 'resolution form' rather than 'algorithm' because we seek generic approaches to: (1) confront both discrete and continuous UCSPs; (2) accommodate broad, heterogeneous classes of constraints and uncertain data; and (3) leverage a range of solving methods from disparate fields. The resolution forms provide templates, showing how to derive closures; we instantiate them for a selection of application domains, and tractable constraint classes. Given the high complexity of finding a closure to a UCSP, we assume independence of parameters when we present tractable instances of the resolution forms, but emphasise that neither the UCSP model nor the resolutions forms themselves require parameter independence. Section 5.5 returns to the issue of approximation.

Briefly, the first resolution form is to transform the UCSP: we find and then solve an equivalent certain CSP with respect to the solution sought. The second form, enumeration, applies when there are a finite number of realisations. Each realisation gives rise to a certain CSP, which we solve, and a closure is then formed from all the solutions to the satisfiable CSPs.

5.2.1 *Comparing Constraints.* The foundation of both resolution forms is reasoning about uncertain constraints in terms of certain constraints. The idea behind such reasoning is that it abstracts us away from the computation domain (i.e. the types of constraints and their representation) and the uncertainty sets (i.e. the nature of the parameters, discrete or continuous, and the representation of the uncertainty sets). We thus require a notion of equivalence between the solutions of uncertain and certain constraints. The central idea — allowing us to compare the solutions sets of UCSPs (which we seek) and CSPs (which we use to describe a closure) — is the observation that uncertain constraints, with suitable operations, form an algebraic structure. The ordering relation and operations in this structure will enable us to compare sets of constraints.

Conjunction, disjunction and implication are defined for certain constraints in the usual way. We extend those definitions to uncertain constraints and define implication between uncertain and certain constraints.

PROPOSITION 5.1. *Let $\mathbb{C}$ be the set of all constraints, certain and uncertain that define a UCSP. With conjunction and disjunction as meet and join, $\mathbb{C}$ is a distributive lattice. With logical implication of constraints, $\mathbb{C}$ has a natural partial order. Let $\widehat{\mathbb{C}} \subset \mathbb{C}$ be the subset of certain constraints; then $\widehat{\mathbb{C}}$ forms a sublattice. In $\mathbb{C}$ and $\widehat{\mathbb{C}}$, the top $\top$ is the always-true* universal *constraint, the bottom $\perp$ the always-false* empty *constraint.*[8]   $\square$

Crucially, the notion of implication in $\mathbb{C}$ takes into account potential realisations of the constraints. We say that an assignment to its variables *satisfies* an uncertain constraint if it satisfies *at least one* of its realised constraints. This ensures that, whatever the true realisation of the parameters, no solution to that realised CSP is a priori excluded.

---

[8]To be precise, elements of $\mathbb{C}$, and top and bottom in particular, are equivalence classes [·] under constraint implication. That is, the top is in fact the class [$\top$]. We will not need to account further for this subtlety.

Working within this algebraic structure, we can encapsulate the solving process using mappings from $\mathbb{C}$ to itself. Reliable solutions in the framework will be guaranteed by properties of the mappings, and correctness can be proved independently of the constraint and data representations. Second, knowledge of the uncertain data can be seen in terms of a *subsumed-by* order on constraints, e.g. should we learn more about the data, the revised closure will be subsumed by the old.

For certain constraints, we have the subsumed-by order [Tsang 1993]. We extend this order to $\mathbb{C}$ by defining an uncertain constraint $c_2$ to *subsume* a constraint $c_1$ if the complete solution set of $c_2$ contains that of $c_1$:

*Definition* 5.2. Let $\preceq$ be an extension of the subsumed-by partial order on $\widehat{\mathbb{C}}$ to $\mathbb{C}$ such that $c_2 \in \mathbb{C}$ *subsumes* $c_1 \in \mathbb{C}$, written $c_1 \preceq c_2$, iff $\mathrm{Cl}(c_2)$ subsumes $\mathrm{Cl}(c_1)$.    □

This partial order is well-defined because $\mathrm{Cl}(c)$ is always a certain constraint. We will write $\preceq$ for the subsumed-by orders both on $\widehat{\mathbb{C}}$ alone and on $\mathbb{C}$, differentiating the two only if required. For UCSPs, $\preceq$ is a formal definition of 'enclosing solution'; it also provides the notion of equivalence we seek:

*Definition* 5.3. Let $c_1, c_2 \in \mathbb{C}$ be uncertain constraints. We say that $c_1$ and $c_2$ are *equivalent* iff $c_1 =_{\preceq} c_2$, i.e. iff $c_1 \preceq c_2$ and $c_2 \preceq c_1$.    □

EXAMPLE 5.2. *Consider the constraints $c_1$: $X > \{2, 3, 4\}$ and $\hat{c}_1$: $X > 2$. We have both $c_1 \preceq \hat{c}_1$ and $\hat{c}_1 \preceq c_1$. Thus $c_1$ and $\hat{c}_1$ are equivalent under $\preceq$ (note $\hat{c}_1$ is the full closure of $c_1$): they describe the same set of potential solutions for $X$.*    □

Thus with $\preceq$ on $\mathbb{C}$ we have a means of deciding when two constraints are equivalent, in terms of their solution sets. We can now give the operational semantics for UCSP solving. Note that the notion of equivalence defined here is for satisfaction problems. Our results extend directly to optimisation UCSPs, once we define the equivalence notion to include a notion of optimality as well as reliability [Yorke-Smith and Gervet 2005].

5.2.2 *Solution Operators.* A classical CSP is solved by propagation and search: one calculates the fixed-point of some local consistency operators and (if necessary) explores the search space. Until we come to a specific application domain, the specific methods used to solve CSPs are not relevant: the essential point is to guarantee correct and tight inference.

Since we are not committed to techniques from one paradigm, we choose to abstract from the algorithms of CSP solving and encapsulate the operational semantics by a *solution operator*. We have seen that a CSP $\widehat{P}$ is described by an element of a certain constraint lattice $\widehat{\mathbb{C}}$ and, likewise, so is every solution to $\widehat{P}$. Hence we define a solution operator as a map from $\widehat{\mathbb{C}}$ to itself that provides the conjunction of a set of solutions to a CSP $\widehat{P}$. The conjunction may be empty, indeed must be if $\widehat{P}$ is inconsistent. Clearly, a solution operator obeys the properties expected of CSP solving [Apt 1999]:

*Definition* 5.4. Let $\widehat{P}$ be a certain CSP. Let $\phi : \widehat{\mathbb{C}} \to \widehat{\mathbb{C}}$ be a map such that $\phi(\mathcal{C})$ describes a set of solutions to $\widehat{P}$. If $\phi$ obeys:

   1.   Contraction      The final state is a subset of the initial state: $\phi(\mathcal{C}) \preceq \mathcal{C}$

   2.   Monotone         Order respected: $\mathcal{C}_1 \preceq \mathcal{C}_2 \implies \phi(\mathcal{C}_1) \preceq \phi(\mathcal{C}_2)$

   3.   Idempotence     Further application of $\phi$ yields no further solutions

then we say that $\phi$ is a *solution operator*. If further $\phi(\mathcal{C})$ describes the set of all solutions to $\widehat{P}$, we say $\phi$ is *complete* for $\widehat{P}$. □

EXAMPLE 5.3. *Consider two solution operators for finite domain CSPs. Let $\phi_1$ be a map that corresponds to naive backtrack search. If we insist that the whole search tree be explored, then $\phi_1$ will give all solutions; this makes it complete. Let $\phi_2$ be a map that corresponds to search with forward checking. Operationally, $\phi_2$ would be expected to be more efficient than $\phi_1$, i.e. find solutions in less time, but declaratively the two operators serve the same purpose for the class of CSPs.* □

Similarly, a solution operator for an uncertain CSP is a map that yields a closure when given a UCSP $P$. A *complete* uncertain solution operator is one that yields the full closure $\mathrm{Cl}(P)$. An uncertain solution operator is formally defined as a mapping from $\mathbb{C}$ to $\widehat{\mathbb{C}}$:

*Definition* 5.5. Let $P$ be a UCSP. An *uncertain solution operator* is a map $\rho : \mathbb{C} \to \widehat{\mathbb{C}}$ such that $\rho(\mathcal{C}) \preceq \mathrm{Cl}(P)$. An uncertain solution operator $\rho$ must obey the contraction, monotone and idempotence properties. If further $\rho(\mathcal{C}) = \mathrm{Cl}(P)$, we say $\rho$ is *complete* for $P$. □

EXAMPLE 5.4. *Over the computation domain of Example 3.1, consider the map $\rho$ that takes the (disjunction of the) constraints of the UCSP $P$ in the example, and returns constraint $X = 3 \wedge Y = 1$. $\rho$ is an uncertain solution operator; it is not complete because it does not describe all of $\mathrm{Cl}(P)$.* □

### 5.3 Transformation Resolution Form

Transformation to an equivalent certain CSP is one way to build an uncertain solution operator; enumeration is another. The idea behind transforming the UCSP is simple: whereas it might be difficult to directly solve the UCSP, there exist efficient means of solving classical CSPs (besides various extensions of them). We would like to take advantage of these powerful solving techniques.

We saw the transformation resolution form in the earlier NTAP case study. In general, it consists of the same two steps. The first step is to transform the UCSP $P$ to an equivalent certain CSP $\tau(P)$. The second step is to find the sought closure of this latter CSP. The practical issues related to this approach are thus also twofold: finding a CSP equivalent to the UCSP, i.e. one whose complete solution set coincides with the sought closure to the original problem; and then solving it efficiently. We achieve the first part by seeking a transformation operator from UCSP to CSP that satisfies specific properties. We achieve the second part by using existing techniques appropriate to the computation domain.

In the scope of this article, we concentrate the discussion of this resolution form to the case of the full closure.

The equivalent CSP is found using a map from the space of UCSPs to the space of CSPs, which, viewed as constraints, corresponds to a map from $\mathbb{C}$ to $\widehat{\mathbb{C}}$. We call a suitable map a *certain equivalence transform* (CET):

*Definition* 5.6. A lattice homomorphism $\tau : \mathbb{C} \to \widehat{\mathbb{C}}$ is a *certain equivalence transform* if it preserves certainty and is monotone and idempotent:

| | | | |
|---|---|---|---|
| 1. | Preserves certainty | $\tau(\hat{c}) = \hat{c}$ | $\forall \hat{c} \in \widehat{\mathbb{C}}$ |
| 2. | Increasing | $c \preceq \tau(c)$ | $\forall c \in \mathbb{C}$ |
| 3. | Monotone | $c_1 \preceq c_2 \Rightarrow \tau(c_1) \preceq \tau(c_2)$ | $\forall c_1, c_2 \in \mathbb{C}$ |
| 4. | Idempotence[9] | $\tau^2(c) = \tau(c)$ | $\forall c \in \mathbb{C}$ |

These properties of $\tau$ are enough to prove that a certain CSP is found and to guarantee the correctness of an uncertain solution operator based on $\tau$. Together, the properties of a CET ensure that the complete solution set of the equivalent CSP contains the full closure to the original problem.

Further, if the full closure to the original problem contains the complete solution set to the equivalent CSP, then $\tau$ is a *tight* CET and the solutions sets of $P$ and $\tau(P)$ are equivalent.

Proposition 5.7 sums up the result: an uncertain solution operator $\rho$ can be defined as the composition of a tight CET $\tau$ and a certain solution operator $\phi$. We can think of $\rho$ as first finding an equivalent certain problem, then applying a certain solution operator to it.

A tight CET thus gives a correct and tight closure. If $\tau$ is a non-tight CET, we can obtain only an outer approximation to the full closure, as shown earlier in Figure 1. In practice, such approximations to the closure are often valuable for their reduced computational or representational cost, as Section 5.5 will discuss.

PROPOSITION 5.7. *Let $P$ be a UCSP. If $\tau$ is a tight CET and $\phi$ is a certain solution operator complete for $\tau(\mathcal{C})$, then $\rho = \phi \circ \tau$ is an uncertain solution operator, complete for $P$.*

PROOF. Let $\langle \mathcal{V}, \mathcal{D}, \Lambda, \mathcal{U}, \mathcal{C} \rangle = P$ and $\mathcal{C} = \bigwedge_i c_i$. Suppose $\tau$ is a tight CET, and $\phi$ is a complete solution operator for $\tau(\mathcal{C})$. We show that $\rho = \phi \circ \tau$ satisfies Definition 5.5. First, since $\tau$ maps $\mathbb{C} \to \widehat{\mathbb{C}}$ and $\phi$ maps $\widehat{\mathbb{C}} \to \widehat{\mathbb{C}}$, then $\rho$ maps $\mathbb{C} \to \widehat{\mathbb{C}}$.

We show that the set of all solutions to the equivalent CSP coincides with the full closure of the original UCSP. That is, $\rho(P) = \text{Cl}(P)$. We prove correctness, then tightness.

From the closure properties of $\tau$, we have for all $c \in \mathbb{C}$, $c \preceq \tau(c) \Rightarrow \text{Cl}(c) \preceq_{\widehat{\mathbb{C}}} \text{Cl}(\tau(c))$ (where $\preceq_{\widehat{\mathbb{C}}}$ is the subsumed-by order over $\widehat{\mathbb{C}}$; we will omit the subscript from now on). This holds for all $c \in \mathbb{C}$; in particular for $\mathcal{C}$. Thus we have $\text{Cl}(\mathcal{C}) \preceq \text{Cl}(\tau(\mathcal{C}))$. By the properties of $\tau$, $\tau(\mathcal{C})$ is a certain constraint. Since $\phi$ is complete for $\tau(\mathcal{C})$, $\text{Cl}(\tau(\mathcal{C})) = \phi(\tau(\mathcal{C}))$ by construction. This gives $\text{Cl}(\mathcal{C}) \preceq \phi(\tau(\mathcal{C}))$.

The converse, $\phi(\tau(\mathcal{C})) \preceq \text{Cl}(\mathcal{C})$, follows at once from the tightness of $\tau$. Hence, since $P$ is given by the conjunction of its constraints $\mathcal{C}$, we have $\rho(P) = \text{Cl}(P)$.

---

[9]Observe that property 4 (idempotence) follows from property 1 (preservation of certainty).

It remains to prove $\rho$ satisfies the contraction, monotone and idempotence properties. The latter two (monotone and idempotence) follow at once from the same properties of $\tau$ and $\phi$, and because $\tau$ preserves certainty.

Contraction follows because $\tau$ is tight, $\phi$ is idempotent, and because of the definition of $\preceq$:

$$
\begin{aligned}
\mathrm{Cl}(\rho(\mathcal{C})) = \mathrm{Cl}(\phi \circ \tau(\mathcal{C})) \quad & \text{since } \rho(P) = \mathrm{Cl}(P), \\
& \text{and } \mathrm{Cl}(P) = \phi \circ \tau(P) \text{ by above} \\
= \phi \circ \phi \circ \tau(\mathcal{C}) \quad & \phi \text{ complete solution operator,} \\
& \text{and } \phi \circ \tau(\mathcal{C}) \in \widehat{\mathbb{C}} \\
= \phi \circ \tau(\mathcal{C}) \quad & \phi \text{ idempotent} \\
\subseteq \mathrm{Cl}(\mathcal{C}) \quad & \tau \text{ tight}
\end{aligned}
$$

whence $\rho(\mathcal{C}) \preceq \mathcal{C}$ by the definition of $\preceq$.  □

EXAMPLE 5.5. *Consider linear arithmetic constraints over two variables $V_1$ and $V_2$ with domains in $\mathbb{R}^+$. Suppose uncertain constraints of the form $\mathbf{a_1}V_1 + \mathbf{a_2}V_2 \leq \mathbf{a_3}$, where $\mathbf{a_i} = [\underline{\mathbf{a_i}}, \overline{\mathbf{a_i}}]$ are real, closed intervals. Then consider an operator that transforms each constraint separately in the following way:*

$$\mathbf{a_1}V_1 + \mathbf{a_2}V_2 \leq \mathbf{a_3} \rightarrow \underline{\mathbf{a_1}}V_1 + \underline{\mathbf{a_2}}V_2 \leq \overline{\mathbf{a_3}} \tag{13}$$

*This CET $\tau_{ils}$ for the full closure was presented in the NTAP case study. We now prove that it actually defines a tight CET. Note that in Section 4, from the equivalent CSP $\tau_{ils}(\mathcal{C})$ we computed a projection of the full closure, i.e. the final outcome for the problem was an approximation (correct but not tight) of the full closure, despite the CET being tight.*

PROOF. *By construction, the transformation given in Section 4 defines a mapping from a UCSP $P$ to a CSP $P'$. It maps the conjunction of constraints $\mathcal{C} = \bigwedge_i c_i \in \mathbb{C}$ to a conjunction of constraints $\tau_{ils}(\mathcal{C}) = \bigwedge_i \tau_{ils}(c_i) \in \widehat{\mathbb{C}}$.*

*To show that this map is a CET for the full closure, we must show it satisfies the properties of Definition 5.6 and is increasing. Observe that $\tau_{ils}$ leaves a certain constraint unchanged; thus we have preservation of certainty (and so idempotence also). The increasing and monotone properties relate to the interaction of $\tau_{ils}$ with the subsumed-by order $\preceq$. They follow from the statement of the last theorem. Proposition 4.3 states the equivalence of $\tau_{ils}(P)$ and $P$ in terms of complete solution sets. Thus, if $A'x \leq b'$ is the transformed system corresponding to a single constraint $c \in \mathbb{C}$, then $\mathrm{Cl}(c) \preceq_{\widehat{\mathbb{C}}} \Sigma(A', b') \iff \mathrm{Cl}(c) \preceq_{\widehat{\mathbb{C}}} \mathrm{Cl}(\tau_{ils}(c)) \iff c \preceq \tau_{ils}(c)$. Monotonicity is similar.*  □

The main strength of the transformation resolution form is that only one deterministic CSP needs to be solved to find the closure, and existing algorithms can be leveraged to provide a practical algorithm for deriving the sought closure from the equivalent CSP. However, a suitable transformation does not necessarily exist for every UCSP instance or constraint class. We outline existing suitable constraint classes and present a new class based on properties on uncertain constraints.

5.3.1 *Practical Instances of Transformation.* We would like to have a 'black-box' process to derive the sought closure, given an input UCSP. As a step towards this, we present existing and identify new tractable instances of UCSPs suitable to the transformation resolution form that satisfy the operational semantics we defined. For each UCSP instance, we show how the transformation resolution form can be instantiated to a polynomial time algorithm. We specify (1) how to transform the problem by a CET, and (2) how to solve the equivalent CSP effectively.

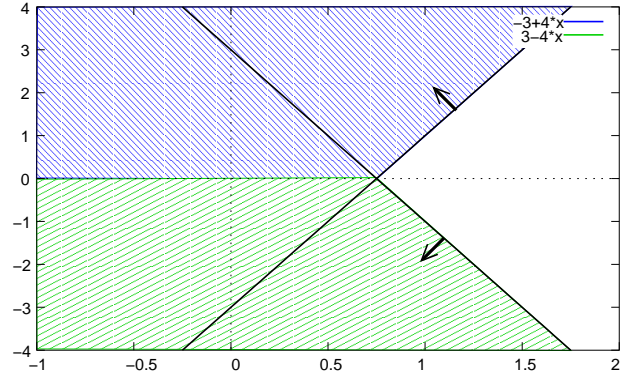

Fig. 6. Polynomial inequality CET described by two constraints. The upper and lower shaded areas denote the two constraints; it can be verified that the solution to every realised constraint lies in one of them.

$\mathcal{D} = \mathbb{R}^{\mathbf{m}}$, $\mathcal{U} = \mathbb{R}^{\boldsymbol{\ell}}$, *and* $\mathcal{C} = \{$*n-ary linear, arithmetic constraints*$\}$.
Example constraint: $[16, 20.5]X + Y - Z \leq [-2, 5]$.

A continuous UCSP with n-ary linear arithmetic constraints can be solved efficiently by transformation. When the constraints are restricted to polynomial inequalities, a tight CET is obtained by examining the bounds on the parameters (a description and proof for the case of non-negative variables were given in Section 4). Note the result holds for both discrete and continuous UCSPs. The resulting equivalent CSP, a system of polynomial inequalities over the reals, is easily solvable with classical computational algebra, or with linear programming.

EXAMPLE 5.6. *Consider the UCSP with real variables* $X, Y \in \mathbb{R}$ *(note the variable domains here are not non-negative: the UCSP is an ILS but is not a POLI system), and the single constraint:*

$$\lambda_1 X + \lambda_2 Y \leq \mu \tag{14}$$

*where the uncertainty sets for the parameters are real intervals:* $U_{\lambda_1} = [4, 5]$, $U_{\lambda_2} = [-1, 1]$, $U_\mu = [2, 3]$. *The CET described above transforms* (14) *into a classical constraint according to:*

$$\begin{cases} Y \geq -3 + 4X & \text{if } Y \geq 0 \\ Y < 3 - 4X & \text{if } Y < 0 \end{cases} \tag{15}$$

*The transformed constraint is depicted in Figure 6.* ☐

$\mathcal{D} = \mathbb{R}^{\mathbf{m}}/\mathbb{Z}^{\mathbf{m}}$, $\mathcal{U} = \mathbb{R}^{\boldsymbol{\ell}}/\mathbb{Z}^{\boldsymbol{\ell}}$, *and* $\mathcal{C} = \{$*simple temporal constraints*$\}$.
Example constraint: $|Y - X| \leq [10, 20]$.

Simple Temporal Problems with Uncertainty (STPUs) [Vidal and Fargier 1999] feature simple temporal constraints over continuous or discrete time-point variables, some of which may have uncontrollable durations. At a conceptual level, existing algorithms to solve a STPU (e.g. fastDCcheck [Morris 2006]) can been seen as instances of the transformation resolution form. They transform $P$ into a STP $P'$ which describes the minimal network for the time-points. Since $P'$ is already in solved form, $\phi$ is trivial. $P'$ represents the full closure of the STPU, since every solution that could occur under one or more realisations is in the minimal network; in fact, since only solutions are in the network, it is a tight representation of the closure. The complexity of the complete process is polynomial in the size of the temporal network [Morris 2006].

Uncertain simple temporal constraints are instances of a more general constraint class we call (binary) *parameter monotone constraints* with one parameter. A STPU with independent parameters is a UCSP with constraints of this form.

$\mathcal{D} = \mathbb{Z}^{\mathbf{m}}$, $\mathcal{U} = \mathbb{Z}^{\boldsymbol{\ell}}$, *and* $\mathcal{C} = \{$*n-ary parameter monotone constraints*$\}$.
Example constraint: $\{-3, 0, 3\}X \le 2Z + \{2, 3, 5\}$.

With parameter monotone constraint we extend the concept of a monotone (certain) constraint [van Beek and Dechter 1995] to uncertain constraints, by defining a structural property of *parameter monotone*. We define the simple case of a binary constraint but it can be generalised to *n*-ary constraints [Yorke-Smith 2004].

*Definition* 5.8. Let $c \in \mathbb{C}$ be an uncertain constraint with two discrete variables and one discrete parameter. If there exists an ordering of the uncertainty set of the parameter such that:

$$\forall \lambda \in U, \ \ \hat{c}(x, y; \lambda) \Rightarrow \hat{c}(x, y; \lambda') \ \forall \lambda' > \lambda \tag{16}$$

we say that $c$ is *parameter monotone.* $\square$

In the case of binary constraints, there exists a simple test for the parameter monotone property. The CET for parameter monotone constraints works on the extremal values of the uncertainty sets. When the constraints are binary, it transforms a parameter monotone, basic, monotone constraint into a binary monotone constraint. The complete solution set of the latter can be found in linear time (in the number of variables) with 2D integer hull algorithms [Harvey 1999]. For example, $\{-3, 0, 3\}X \le 2Z + \{2, 3, 5\}$ is transformed to $-3X \le 2Z + 5$, for $X \ge 0$.

*Remark* 5.9. We highlight that the transformation resolution form extends to optimisation problems. RO seeks transformation of well-structured optimisation UCSPs (with complex forms of uncertainty sets) to a robust counterpart, in order to inherit the computational complexity of the equivalent optimisation CSP (CSOP) [Hoffman 2000]. The robust counterpart is an equivalent CSOP with respect to the optimality of the solution sought.

We also note that the transformation resolution form applies to other types of closure beyond the full closure. For instance, in Example 5.6, if the user desires the robust closure rather than the full closure, the CET obtains the intersection rather than the union of the transformed constraints:

$$\begin{cases} Y < 3 - 4X & \text{if } Y \geq 0 \\ Y \geq -3 + 4X & \text{if } Y < 0 \end{cases} \tag{17}$$

### 5.4 Enumeration Resolution Form

Seeking a CET is possible when the uncertain problem is well-structured. It has the main advantage of benefiting from the computational complexity of the equivalent deterministic CSP. However, depending on the constraint class, it might not always be possible to find a CET, and depending on the problem, the user might want to know which solution supports which realisation. This support information is not guaranteed to be preserved by a CET. A second and complementary means to derive closures is by enumeration. Clearly, as an essentially exhaustive technique, enumeration requires operationally there be only finitely-many, $M < \infty$, realisations of the data. We generate and solve each realised CSP, forming the closure from the solutions to all the good realisations.

Contrasted with transformation, the cost of enumerating and solving $M$ possibly similar CSPs grows with $M$, which can be exponential in the size of the UCSP. This said, there are three mitigating factors. First, in a given computation domain, it is important to exploit knowledge of the structure of the realised problems. Second, it may be possible to solve one realised CSP based on the experience of solving the previous ones, using specific dynamic CSP techniques (see the survey [Verfaillie and Jussien 2005]). Third, the solving of the realised CSPs can in principle be parallelised.

Straight-forward enumeration can be polynomial in space, if each realised CSP is in PSPACE, but is usually exponential in time. For practical use, the challenge (as hinted earlier) is to provide efficient operational semantics in specific cases. Two ways to reduce the time complexity are (1) solve each realised CSP more efficiently (as noted above); and (2) solve only a subset of the realised CSPs.

#### 5.4.1 Instances of the Enumeration Form for Different Closures.

We sketch how existing solution methods can be leveraged to provide algorithms for deriving different types of closures.

*Full closure with continuous data.* If reals are finitely represented (e.g. as in floating point arithmetic), enumeration is commonly used to derive the complete solution set of NCSPs. The branch-and-prune methodology [Benhamou 1995] is such an instance allowing us to solve a UCSP with continuous data and variables, provided all the constraints are negable (e.g. $[15, 25]X + \sin(Y) \neq XZ^2 + 1$). It has been extended to classes of constraints including one equality [Ratschan 2003].

Besides branch-and-prune, other instances exist such as symbolic quantifier elimination methods (expensive but exact), and Monte Carlo estimation with application to arbitrary n-ary polynomial inequalities over the reals.

*Full closure with discrete data.* The general CSP algorithm *constructive disjunction* (CD) [Van Hentenryck et al. 1998] can be used seen as an instance of the enumeration form to derive the full closure. Indeed, if we consider a UCSP $P$ as a disjunction of its realised CSPs, $\bigvee_i \widehat{P_i}$, then $\text{Cl}(P)$ is a constraint implied by the disjunction. In constructive disjunction one eliminates all domain values not supported in at least one of the disjuncts (i.e. not supported by at least one realisation). This is exactly what is required to derive the full closure. The improved algorithm for CD from [Lhomme 2003] can be made incremental by storing supports for each satisfiable realisation. Techniques for discrete QCSPs, such as quantified arc consistency, can also be used as preprocessing steps.

*Most robust solution and covering set closures with discrete data.* CSP algorithms extended for mixed CSPs [Fargier et al. 1996] are instances of the enumeration resolution form. These essentially enumerative algorithms apply directly to the discrete data case of UCSPs over finite domains, with the uncertainty represented by subsets of $\mathbb{Z}$. They are efficient when there is at most one parameter per constraint and the data is independent.

There are two algorithms for mixed CSP solving, depending on whether further knowledge about the realisations will be acquired. If not, the *no observability* case, a most robust solution closure (in our terminology) is sought by branch-and-bound search with forward checking. If instead the true state of the world will become known to us in the future, the *full observability* case, a covering set closure (in our terminology) is sought by decomposing the space of realisations.

## 5.5   Approximation

Studies in OR demonstrate that approximation can be the only tractable way to solve certain problems, and that in practice the information obtained often suffices. The main reasons we consider approximations in the certainty closure framework are at heart operational. Seeking tractable models can lead us to trade-off tightness of the closure and efficiency in deriving or representing it. Approximation must not impair correctness but may forgo tightness: for the full closure, for instance, it must not exclude any potential solutions but may include unsupported 'solutions'.

To reduce the computational complexity, we can choose to approximate the model, the solution sought, or its representation (approximating the model leads to approximating the solution but not necessarily the reverse) [Yorke-Smith 2004]. First, as in the NTAP case study of Section 4, we can approximate the UCSP model. For the NTAP, we made an approximation by assuming independence of the traffic splitting parameters, i.e. we omitted some constraints from the UCSP. Second, we can seek an outer approximation of the closure, i.e. deliberately derive a non-tight set. This is a common choice in reliable computation. For instance, we can approximate the uncertain solution operator by an outer-enclosing operator, such as that derived from a non-tight CET (recall Section 5.3). Third, we can choose a constraint class that can express only an inexact representation of the closure [Maher 2003]. For the NTAP, we used the class of axis-parallel hyperboxes rather than general affine constraints (recall Figure 3), i.e. we projected the full closure onto the variables.

## 6.   CONCLUSION

In this paper we defined the *certainty closure framework*, by introducing a generic UCSP model to represent constraint problems with incomplete and erroneous data, and a closure as the sought outcome. Our framework demonstrates how existing models and methods from different fields such as reliable computation and op-erational research can be brought together as a tractable means towards reliable reasoning in the presence of incomplete and erroneous data. It constitutes a first step towards automating the solving of UCSPs using a generic, non-probabilistic model that encloses data uncertainty within convex uncertainty sets. We have given specific instances of two resolutions forms, and defined new constraint classes suitable for the transformation resolution form to derive the full closure.

Summarised, the benefits of the certainty closure framework include:

—*Generic*. The framework is defined at an abstract level. Thus it is widely applicable across diverse application domains. It can be instantiated to a concrete form in each case.
—*Reliable*. The closure paradigm ensures that each uncertain CSP is a reliable model of the ill-defined constraint problems. Properties of the framework guarantee that reliable solutions are obtained from the model.
—*Practical*. The approach seeks efficient algorithms to solve the model, as instances of the resolution forms.
—*Unifying*. Ideas and methods from different fields are brought together by the framework: described in a common language and embedded in the CP formalism.

In the Network Traffic Analysis Problem case study, we addressed a real-life problem in networking, driven by the user's requirement for reliable bounds on traffic flows. Solving the NTAP as a UCSP for its full closure produced guaranteed bounds, at a computational cost comparable to the initial deterministic, data correction approach. In terms of insights gained, we could diagnose the lack of reliability of the initial approach, and could provide informed feedback to the end user regarding the network behaviour.

*Future work.* This paper has concentrated on developing and illustrating the framework, and in particular the resolution forms, for the full closure. Section 5.1 listed other types of closures to a UCSP. Part of our future is to seek tractable instances for such alternative closures. One aspect related to tractability is approximation. In addition, we are interested in identifying constraint classes and uncertainty set representations for which accounting for data dependency does not render the problem intractable.

Regarding the NTAP application, our implementation in ECL$^i$PS$^e$ automatically transformed the initial UCSP model into an equivalent certain CSP, then solved by LP. The next challenge is to build a system that detects the UCSP problem structure (constraint and data types) and identifies the most effective instance of a resolution form, based first on the instances presented in this paper.

The UCSP model is extensible to optimisation problems with various optimisation criteria. We recently introduced the *uncertain CSOP* model [Yorke-Smith and Gervet 2005], which opens many fruitful research directions, including defining a suitable notion of optimality with respect to a closure or set of closures, adapting work found in robust optimisation.

## ACKNOWLEDGMENTS

The authors thank many for helpful discussions during the development of this work, in particular our former colleagues at IC–Parc. We are grateful to the reviewers for their suggestions that substantially aided the organisation of this paper. This work was partially supported by the EPSRC under grant GR/N64373/01.

...